\documentclass[10pt,journal,compsoc]{IEEEtran}
\ifCLASSINFOpdf
  \usepackage[pdftex]{graphicx}
\else
  \usepackage[dvips]{graphicx}
\fi
\usepackage{amsmath}
\usepackage{algorithmic}

\usepackage{array}

\ifCLASSOPTIONcompsoc
  \usepackage[caption=false,font=footnotesize,labelfont=sf,textfont=sf]{subfig}
\else
  \usepackage[caption=false,font=footnotesize]{subfig}
\fi
\usepackage{url}

\usepackage{multirow}
\usepackage{hyperref}
\usepackage{amsmath,amsthm,amssymb,amsfonts,mathabx} 
\usepackage{multibib}
\usepackage{ragged2e}
\usepackage{array}

\newtheorem{theorem}{Theorem}

\begin{document}
%
\title{Heatmap Regression via Randomized Rounding}
%
%
%
%

\author{Baosheng~Yu~and~Dacheng~Tao,~\IEEEmembership{Fellow,~IEEE}
\IEEEcompsocitemizethanks{\IEEEcompsocthanksitem Baosheng Yu is with The University of Sydney, Australia. E-mail: baosheng.yu@sydney.edu.au.
\IEEEcompsocthanksitem  Dacheng Tao is with JD Explore Academy, China and The University of Sydney, Australia. Email: dacheng.tao@gmail.com.
\IEEEcompsocthanksitem  Corresponding author: Dacheng Tao.}
\thanks{1057-7149 © 2018 IEEE. Personal use of this material is permitted. Permission from IEEE must be obtained for all other uses, in any current or future media, including reprinting/republishing this material for advertising or promotional purposes, creating new collective works, for resale or redistribution to servers or lists, or reuse of any copyrighted component of this work in other works.}}

\markboth{IEEE TRANSACTIONS ON PATTERN ANALYSIS AND MACHINE INTELLIGENCE}%
{Shell \MakeLowercase{\textit{et al.}}: Bare Demo of IEEEtran.cls for Computer Society Journals}


\IEEEtitleabstractindextext{%
\justify
\begin{abstract}
Heatmap regression has become the mainstream methodology for deep learning-based semantic landmark localization, including in facial landmark localization and human pose estimation. Though heatmap regression is robust to large variations in pose, illumination, and occlusion in unconstrained settings, it usually suffers from a sub-pixel localization problem. Specifically, considering that the activation point indices in heatmaps  are always integers, quantization error thus appears when using heatmaps as the representation of numerical coordinates. Previous methods to overcome the sub-pixel localization problem usually rely on high-resolution heatmaps. As a result, there is always a trade-off between achieving localization accuracy and computational cost, where the computational complexity of heatmap regression depends on the heatmap resolution in a quadratic manner. In this paper, we formally analyze the quantization error of vanilla heatmap regression and propose a simple yet effective quantization system to address the sub-pixel localization problem. The proposed quantization system induced by the \textit{randomized rounding operation} 1) encodes the fractional part of numerical coordinates into the ground truth heatmap using a probabilistic approach during training; and 2) decodes the predicted numerical coordinates from a set of activation points during testing. We prove that the proposed quantization system for heatmap regression is unbiased and lossless. Experimental results on popular facial landmark localization datasets (WFLW, 300W, COFW, and AFLW) and human pose estimation datasets (MPII and COCO) demonstrate the effectiveness of the proposed method for efficient and accurate semantic landmark localization. Code is available at~\url{http://github.com/baoshengyu/H3R}.
\end{abstract}
}

\maketitle

\IEEEdisplaynontitleabstractindextext

%
\IEEEpeerreviewmaketitle

\IEEEraisesectionheading{\section{Introduction}\label{sec:introduction}}

\IEEEPARstart{S}{emantic} landmarks are sets of points or pixels in images containing rich semantic information. They reflect the intrinsic structure or shape of objects such as human faces~\cite{sun2013deep,bulat2017far}, hands~\cite{sinha2016deephand,iqbal2018hand}, bodies~\cite{toshev2014deeppose,newell2016stacked}, and household objects~\cite{ashutosh2008robotic}. Semantic landmark localization is fundamental in computer and robot vision~\cite{sun2014deep,thies2016face2face,ashutosh2008robotic,wang2019densefusion}. For example, semantic landmark localization can be used to register correspondences between spatial positions and semantics (semantic alignment), which is extremely useful in visual recognition tasks such as face recognition~\cite{sun2014deep,deng2018arcface} and person re-identification~\cite{zhao2017spindle,zheng2019pose}. Therefore, robust and efficient semantic landmark localization is extremely important in applications requiring accurate semantic landmarks including robotic grasping~\cite{ashutosh2008robotic,wang2019densefusion} and facial analysis applications such as face makeup~\cite{chen2017makeup,chang2018pairedcyclegan}, animation~\cite{cao20133d,cao2014displaced}, and reenactment~\cite{thies2015real,thies2016face2face}.

\begin{figure}[!ht]
\begin{center}
\includegraphics[width=\linewidth]{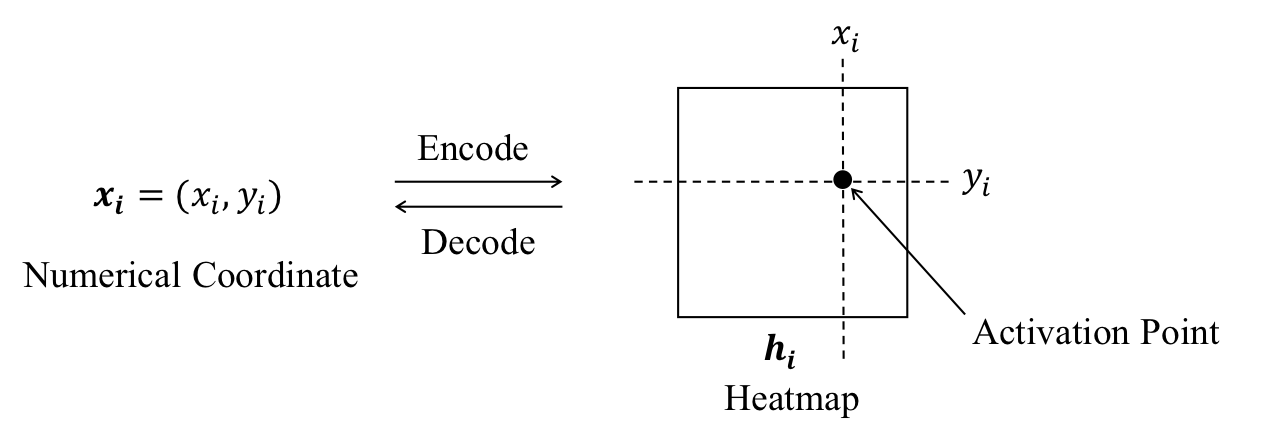}
\caption{An intuitive example of the heatmap representation for the numerical coordinate. Given the numerical coordinate $\boldsymbol{x}_i=(x_i,y_i)$, we then have the corresponding heatmap $\boldsymbol{h}_i$ with the maximum activation point located at the position $(x_i,y_i)$. }
\label{fig:coordinate-heatmap}
\end{center}
\end{figure}

Coordinate regression and heatmap regression are two widely-used methods for deep learning-based semantic landmark localization~\cite{sun2013deep,tompson2014joint}. Rather than directly regressing the numerical coordinate with a fully-connected layer, heatmap-based methods aim to predict the heatmap where the maximum activation point corresponds to the semantic landmark in the input image. An intuitive example of heatmap representation is shown in Fig.~\ref{fig:coordinate-heatmap}. Due to the effective spatial generalization of heatmap representation, heatmap regression method is robust to large variations in pose, illumination, and occlusion in unconstrained settings~\cite{tompson2014joint,nibali2018numerical}. Heatmap regression has performed particularly well in semantic landmark localization tasks including facial landmark detection~\cite{bulat2017far,wang2019adaptive} and human pose estimation~\cite{newell2016stacked,sun2019deep}. Despite this promise, heatmap regression method suffers from an inherent drawback, namely that the indices of the activation points in heatmaps are always integers. Vanilla heatmap-based methods therefore fail to predict the numerical coordinates in sub-pixel precision. Sub-pixel localization is nevertheless important in real-world scenarios with the fractional part of numerical coordinates originating from: 1) the input image being captured either by a low-resolution camera and/or at a relatively large distance; and 2) the heatmap is usually at a much lower resolution than the input image due to the downsampling stride of convolutional neural networks. As a result, low-resolution heatmaps significantly degrade heatmap regression performance. Considering that the computational cost of convolutional neural networks usually depends quadratically on the resolution of the input image or the feature map, there is a trade-off between the localization accuracy and the computational cost for heatmap regression~\cite{newell2016stacked,sun2019high,tai2019towards,li2019rethinking,zhang2019distribution}. Furthermore, the downsampling stride of heatmap is not always equal to the downsampling stride of feature map: given an original image of $512\times512$ pixels, a heatmap regression model with the input size $128\times128$ pixels, and the feature map with a downsampling stride $4$ pixels, we then have the size of heatmap $32\times32$ pixels, i.e., the downsampling stride of heatmap $s=16$ pixels. For simplicity, we do not distinguish between the above two settings to address the quantization error in a unified manner. Unless otherwise mentioned, we refer to $s>1$ as the downsampling stride of the heatmap.

In vanilla heatmap regression, 1) during training, the ground truth numerical coordinates are first quantized to generate the ground truth heatmap; and 2) during testing, the predicted numerical coordinates can be decoded from the maximum activation point in the predicted heatmap. However, typical quantization operations such as \textbf{floor}, \textbf{round}, and \textbf{ceil} discard the fractional part of the ground truth numerical coordinates, making it difficult to reconstruct the fractional part even from the optimal predicted heatmap. This error induced by the transformation between numerical coordinates and heatmap is known as the quantization error. To address the problem of quantization error, here we introduce a new quantization system to form a lossless transformation between the numerical coordinates and the heatmap. In our approach, during training, the proposed quantization system uses a set of activation points, and the fractional part of the numerical coordinate is encoded as the activation probabilities of different activation points. During testing, the fractional part can then be reconstructed according to the activation probabilities of the top $k$ maximum activation points in the predicted heatmap. To achieve this, we introduce a new quantization operation called randomized rounding, or \textbf{random-round}, which is widely used in combinatorial optimization to convert fractional solutions into integer solutions with provable approximation guarantees~\cite{raghavan1987randomized,korte2012combinatorial}. Furthermore, the proposed method can easily be implemented using a few lines of source code, making it a plug-and-play replacement for the quantization system of existing heatmap regression methods.

In this paper, we address the problem of quantization error in heatmap regression. The remainder of the paper is structured as follows. In the preliminaries, we briefly review two typical semantic landmark localization methods, coordinate regression and heatmap regression. In the methods, we first formally introduce the problem of quantization error by decomposing the prediction error into the heatmap error and the quantization error. We then discuss quantization bias in vanilla heatmap regression and prove a tight upper bound on the quantization error in vanilla heatmap regression. To address quantization error, we devise a new quantization system and theoretically prove that the proposed quantization system is unbiased and lossless. We also discuss uncertainty in heatmap prediction as well as the unbiased annotation when forming a robust semantic landmark localization system. In the experimental section, we demonstrate the effectiveness of our proposed method on popular facial landmark detection datasets (WFLW, 300W, COFW, and AFLW) and human pose estimation datasets (MPII and COCO).

\section{Related Work} \label{sec:related-work}

Semantic landmark localization, which aims to predict the numerical coordinates for a set of pre-defined semantic landmarks in a given image or video, has a variety of applications in computer and robot vision including facial landmark detection~\cite{sun2013deep,cao2014face,bulat2017far}, hand landmark detection~\cite{sinha2016deephand,iqbal2018hand}, human pose estimation~\cite{toshev2014deeppose,newell2016stacked,sun2019deep}, and household object pose estimation~\cite{ashutosh2008robotic,wang2019densefusion}. In this section, we briefly review recent works on coordinate regression and heatmap regression for semantic landmark localization, especially in facial landmark localization applications.

\subsection{Coordinate Regression}

Coordinate regression has been widely and successfully used in semantic landmark localization under constrained settings, where it usually relies on simple yet effective features~\cite{zhu2012face,xiong2013supervised,ren2014face,kazemi2014one}. To improve the performance of coordinate regression for semantic landmark localization in the wild, several methods have been proposed by using cascade refinement~\cite{sun2013deep,burgos2013robust,zhang2014coarse,zhu2015face,carreira2016human,belagiannis2017recurrent}, parametric/non-parametric shape models~\cite{belhumeur2013localizing,cao2014face,zhu2015face,miao2018direct}, multi-task learning~\cite{zhang2014facial,zhang2016joint,ranjan2017hyperface}, and novel loss functions~\cite{girshick2015fast,feng2018wing}.

\subsection{Heatmap Regression}

The success of deep learning has prompted the use of heatmap regression for semantic landmark localization, especially for robust and accurate facial landmark localization~\cite{bulat2017far,bulat2017binarized,merget2018robust,sun2019high} and human pose estimation~\cite{tompson2014joint,pfister2015flowing,newell2016stacked,pishchulin2016deepcut,newell2017associative,yang2017learning,papandreou2018personlab}. Existing heatmap regression methods either rely on large input images or empirical compensations during inference to mitigate the problem of quantization error~\cite{newell2016stacked,wei2016convolutional,cao2017realtime,sun2019deep}. For example, a simple yet effective compensation method known as ``shift a quarter to the second maximum activation point" has been widely used in many state-of-the-art heatmap regression methods~\cite{newell2016stacked,chen2018cascaded,sun2019deep}. 

Several methods have been developed to address the problem of quantization error in three aspects: 1) jointly predicting the heatmap and the offset in a multi-task manner~\cite{papandreou2017towards}; 2) encoding and decoding the fractional part of numerical coordinates via a modulated 2D Gaussian distribution~\cite{tai2019towards,zhang2019distribution}; and 3) exploring differentiable transformations between the heatmap and the numerical coordinates~\cite{sun2018integral,nibali2018numerical,luvizon20182d,luvizon2019human}. Specifically, \cite{tai2019towards} generates the fractional part sensitive ground truth heatmap for video-based face alignment, which is known as fractional heatmap regression. Under the assumption that the predicted heatmap follows a 2D Gaussian distribution, \cite{zhang2019distribution} decodes the fractional part of numerical coordinates from the modulated predicted heatmap. The soft-argmax operation is differentiable~\cite{yi2016lift,levine2016end,thewlis2017unsupervised,luvizon2019human}, and has been intensively explored in human pose estimation~\cite{sun2018integral,nibali2018numerical}.

\section{Preliminaries} \label{sec:preliminary}

In this section, we introduce two widely-used semantic landmark localization methods, coordinate regression and heatmap regression. For simplicity, we use facial landmark detection as an intuitive example.

\textbf{Coordinate Regression.} Given a face image, semantic landmark detection aims to find the numerical coordinates of a set of pre-defined facial landmarks $\boldsymbol{x}_i = \left(x_i, y_i\right)$, where $i=1, 2, \dots, K$, indicate the indices of different facial landmarks (e.g., a set of five pre-defined facial landmarks can be left eye, right eye, nose, left mouth corner, and right mouth corner). It is natural to train a model (e.g., deep neural networks) to directly regress the numerical coordinates of all facial landmarks. The coordinate regression model then can be optimized via a typical regression criterion such as mean squared error (MSE) and mean absolute error (MAE). For the MSE criterion (also the L2 loss), we have
\begin{equation}\label{eq:coodinate-mse}
   \mathcal{L}(\boldsymbol{x}^p_i,\boldsymbol{x}^{g}_i) =  \|\boldsymbol{x}_i^p-\boldsymbol{x}_i^g\|_{2}^2,
\end{equation}
where $\boldsymbol{x}^p_i$ and $\boldsymbol{x}^g_i$ indicate the predicted and the ground truth numerical coordinates, respectively. When using the MAE criterion (also the L1 loss), the loss function $\mathcal{L}$ can be defined in a similar way to \eqref{eq:coodinate-mse}. 

\begin{figure*}[!t]
\begin{center}
\includegraphics[width=\textwidth]{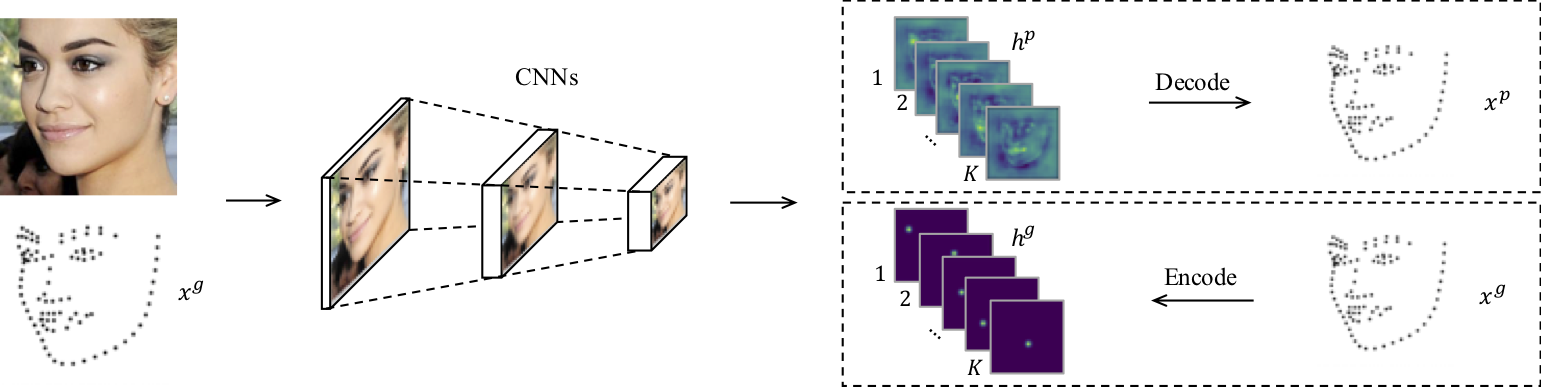}
\caption{The main deep learning-based heatmap regression for semantic landmark localization framework. Specifically, during the inference stage, we decode the predicted heatmaps $\boldsymbol{h}^p=\left(\boldsymbol{h}^p_1, \dots, \boldsymbol{h}^p_K\right)$ to obtain the predicted numerical coordinates $\boldsymbol{x}^p=\left(\boldsymbol{x}^p_1, \dots, \boldsymbol{x}^p_K\right)$; During the training stage, we encode the ground truth numerical coordinates $\boldsymbol{x}^g=\left(\boldsymbol{x}^g_1, \dots, \boldsymbol{x}^g_K\right)$ to generate the ground truth heatmaps $\boldsymbol{h}^g=\left(\boldsymbol{h}^g_1, \dots, \boldsymbol{h}^g_K\right)$.}
\label{fig:framework}
\end{center}
\end{figure*}

\textbf{Heatmap Regression.} Heatmaps (also known as confidence maps) are simple yet effective representations of semantic landmark locations. Given the numerical coordinate $\boldsymbol{x}_i$ for the $i$-th semantic landmark, it then corresponds to a specific heatmap $\boldsymbol{h}_i$ as shown in Fig.~\ref{fig:coordinate-heatmap}. For simplicity, we assume $\boldsymbol{h}_i$ is the same size as the input image in this section and leave the problem of quantization error to the next section. With the heatmap representation, the problem of semantic landmark localization can be translated into heatmap regression via two heatmap subroutines: 1)~\textbf{encode} (from the ground truth numerical coordinate $\boldsymbol{x}_i^g$ to the ground truth heatmap $\boldsymbol{h}_i^g$); and 2) \textbf{decode} (from the predicted heatmap $\boldsymbol{h}_i^p$ to the predicted numerical coordinate $\boldsymbol{x}_i^p$). The main deep learning-based heatmap regression for semantic landmark localization framework is shown in Fig.~\ref{fig:framework}.  
Specifically, during the inference stage, given a predicted heatmap $\boldsymbol{h}_i^p$, the value $\boldsymbol{h}_i^p(\boldsymbol{x}) \in [0,1]$ indicates the confidence score that the $i$-th landmark is located at coordinate $\boldsymbol{x} \in \mathbb{N}^2$. Then, we can decode the predicted numerical coordinate $\boldsymbol{x}_i^p$ from the predicted heatmap $\boldsymbol{h}_i^p$ using the \textbf{argmax} operation, i.e.,
\begin{equation}
    \label{eq:argmax}
    \boldsymbol{x}_i^p = \left(x_i^p, y_i^p\right) \in \underset{\boldsymbol{x}}{\arg\max}~ \left\{ \boldsymbol{h}_i^p(\boldsymbol{x})\right\}.
\end{equation}
Therefore, with the decode operation in~\eqref{eq:argmax}, the problem of semantic landmark localization can be solved by training a deep model to predict heatmap $\boldsymbol{h}_i^p$.

To train a heatmap regression model, the ground truth heatmap $\boldsymbol{h}_i^g$ is indispensable, i.e., we need to encode the ground truth coordinate $\boldsymbol{x}_i^g$ into the ground truth heatmap $\boldsymbol{h}_i^g$.  We introduce two widely-used methods to generate the ground truth heatmap, \textbf{Gaussian heatmap} and \textbf{binary heatmap}, as follows. Given the ground truth coordinate $\boldsymbol{x}_i^g$, the ground truth Gaussian heatmap can be generated by sampling and normalizing from a bivariate normal distribution $\mathcal{N}(\boldsymbol{x}_i^g, \Sigma)$, i.e., the ground truth heatmap $\boldsymbol{h}_i^g$ at location $\boldsymbol{x}\in \mathbb{N}^2$ can be evaluated as 
\begin{equation}
    \label{eq:encode:gaussian}
    \boldsymbol{h}_i^g(\boldsymbol{x}) = \exp\left(-\frac{1}{2}(\boldsymbol{x}-\boldsymbol{x}_i^g)^{\top}\Sigma^{-1}(\boldsymbol{x}-\boldsymbol{x}_i^g)\right),
\end{equation}
where $\Sigma$ is the covariance matrix (a positive semi-definite matrix) and $\sigma > 0$ is the standard deviation in both directions, i.e.,
\begin{equation}
\Sigma = 
\begin{bmatrix}
\sigma^2 & 0 \\
0 & \sigma^2
\end{bmatrix}.
\end{equation}
When $\sigma~\to~0$, the ground truth heatmap can be generated by assigning a positive value at the ground truth numerical coordinate $\boldsymbol{x}_i^g$, i.e.,
\begin{equation}
\label{eq:encode:binary}
\boldsymbol{h}_i^g(\boldsymbol{x}) = 
\begin{cases}
1 & \quad \text{if } \boldsymbol{x}=\boldsymbol{x}_i^g, \\
0 & \quad \text{otherwise}.
\end{cases}
\end{equation}
Specifically, when $\sigma \to 0$, the ground truth heatmap defined in \eqref{eq:encode:binary} is also known as the \textbf{binary heatmap}.

Given the ground truth heatmap, the heatmap regression model then can be optimized using typical pixel-wise regression criteria such as MSE, MAE, or Smooth-L1~\cite{girshick2015fast}. Specifically, for Gaussian heatmaps, the heatmap regression model is usually optimized with the pixel-wise MSE criterion, i.e.,
\begin{equation}\label{eq:heatmap:mse}
   \mathcal{L}(\boldsymbol{h}_i^p,\boldsymbol{h}_i^{g}) = \mathbb{E} \|\boldsymbol{h}_i^p(\boldsymbol{x})-\boldsymbol{h}_i^g (\boldsymbol{x})\|_{2}^2.
\end{equation}
When using the MAE/Smooth-L1 criteria, the loss function can be defined in a similar way to \eqref{eq:heatmap:mse}. For binary heatmap, the heatmap regression model can also be optimized with the pixel-wise cross-entropy criterion, i.e.,
\begin{equation}
\label{eq:loss:binary}
 \mathcal{L}(\boldsymbol{h}^p_i,\boldsymbol{h}^{g}_i) = \mathbb{E} \left(\mathcal{L}_{\text{CE}}\left(\boldsymbol{h}^p_i(\boldsymbol{x}),\boldsymbol{h}^{g}_i(\boldsymbol{x})\right)\right),
 \end{equation}
where $\mathcal{L}_{\text{CE}}$ indicates the cross-entropy criterion with a softmax function as the activation/normalization function.  A comprehensive review of different loss functions for semantic landmark localization is beyond the scope of this paper, but we refer interested readers to \cite{feng2018wing} for descriptions of coordinate regression and \cite{wang2019adaptive} for heatmap regression. Unless otherwise mentioned, we use the MSE criterion for Gaussian heatmap and the cross-entropy criterion for binary heatmap in this paper.

\section{Method}
\label{sec:method}

In this section, we first introduce the quantization system in heatmap regression and then formulate the quantization error in a unified way by correcting the quantization bias in a vanilla quantization system. Lastly, we devise a new quantization system via randomized rounding to address the problem of quantization error.

\subsection{Quantization System}

Heatmap regression for semantic landmark localization usually contains two key components: 1) heatmap prediction; and 2) transformation between the heatmap and the numerical coordinates. The \textbf{quantization system} in heatmap regression is a combination of the \textbf{encode} and \textbf{decode} operations. During training, when the ground truth numerical coordinates $\boldsymbol{x}_i^g$ are floating-point numbers, we then need to calculate a specific Gaussian kernel matrix using \eqref{eq:encode:gaussian} for each landmark, since different numerical coordinates usually have different fractional parts. As a result, it will significantly increase the training loads of the heatmap regression model. For example, given $98$ landmarks per face image, the kernel size $11\times11$, and a mini-batch of $16$ training samples, we then have to run \eqref{eq:encode:gaussian} for $98\times16\times11\times11=189,728$ times in each training iteration. To address this issue, existing heatmap regression methods usually first quantize numerical coordinates into integers, where a standard kernel matrix can then be shared for efficient ground truth heatmap generation~\cite{newell2016stacked,bulat2017far,chen2018cascaded,sun2019high}. However, the above-mentioned existing heatmap regression methods usually suffer from the inherent drawback of failing to encode the fractional part of numerical coordinates. Therefore, how to efficiently encode the fractional information in numerical coordinates still remains challenging. Furthermore, during the inference stage, the predicted numerical coordinates $\boldsymbol{x}_i^p$ obtained by a decode operation in \eqref{eq:argmax} are also integers. As a result, typical heatmap regression methods usually fail to efficiently address the fractional part of the numerical coordinate during both training and inference, resulting in localization error.

To analyze the localization error caused by the quantization system in heatmap regression, we formulate the localization error as the sum of \textbf{heatmap error} and \textbf{quantization error} as follows:
\begin{equation}
\begin{split}
\mathcal{E}_{loc} = \|\boldsymbol{x}_i^p - \boldsymbol{x}_i^g\|_2 &= \|\boldsymbol{x}_i^p  - \boldsymbol{x}_i^{opt} + \boldsymbol{x}_i^{opt} - \boldsymbol{x}_i^g\|_2 \\
&\leq  \underbrace{\|\boldsymbol{x}_i^p  - \boldsymbol{x}_i^{opt}\|_2}_{\textbf{heatmap error}} + \underbrace{\|\boldsymbol{x}_i^{opt} - \boldsymbol{x}_i^g \|_2}_{\textbf{quantization error}},
\end{split}
\end{equation}
where $\boldsymbol{x}_i^{opt}$ indicates the numerical coordinate decoded from the optimal predicted heatmap. Generally, the heatmap error corresponds to the error in heatmap prediction, i.e., $\|\boldsymbol{h}_i^p - \boldsymbol{h}_i^g\|_2$,  and the quantization error indicates the error caused by both the encode and decode operations. If there is no heatmap error, the localization error then all originates from the error of the quantization system, i.e.,
\begin{equation}
\begin{split}
\mathcal{E}_{loc} = \|\boldsymbol{x}_i^p - \boldsymbol{x}_i^g\|_2 = \|\boldsymbol{x}_i^{opt} - \boldsymbol{x}_i^g \|_2.
\end{split}
\end{equation}
The generalizability of deep neural networks for heatmap prediction, i.e., the heatmap error, is beyond the scope of this paper. We do not consider the heatmap error during the analysis of quantization error in this paper.

To obtain integer coordinates for the generation of the ground truth heatmap, typical integer quantization operations such as~\textbf{floor},~\textbf{round}, and~\textbf{ceil} have been widely used in previous heatmap regression methods. To unify the quantization error induced by different integer operations, we first introduce a unified integer quantization operation as follows. Given a downsampling stride $s > 1$ and a threshold $t \in [0, 1]$, the coordinate $x \in \mathbb{N}$ then can be quantized according to its fractional part $\epsilon=x/s - \lfloor x/s \rfloor$, i.e.,
\begin{equation}
\label{eq:quantization}
\boldsymbol{q}(x,s,t) = 
\begin{cases}
\lfloor x/s \rfloor & \quad \text{if }~\epsilon < t, \\
\lfloor x/s \rfloor + 1 & \quad \text{otherwise}.
\end{cases}
\end{equation}
That is, for integer quantization operations~\textbf{floor},~\textbf{round}, and~\textbf{ceil}, we have $t=1.0$, $t=0.5$, and $t=0$, respectively. Furthermore, when the downsampling stride $s > 1$, the decode operation in~\eqref{eq:argmax} then becomes 
\begin{equation}
    \label{eq:argmax:stride}
    \boldsymbol{x}_i^p \in s * \left(\underset{\boldsymbol{x}}{\arg\max}~\left\{\boldsymbol{h}_i^p(\boldsymbol{x})\right\}\right).
\end{equation}
A vanilla quantization system for heatmap regression can then be formed by the encode operation in~\eqref{eq:quantization} and the decode operation in~\eqref{eq:argmax:stride}. When applied to a vector or a matrix, the integer quantization operation defined in~\eqref{eq:quantization} is an element-wise operation.

\begin{figure*}[!ht]
\begin{center}
\centerline{\includegraphics[width=\textwidth]{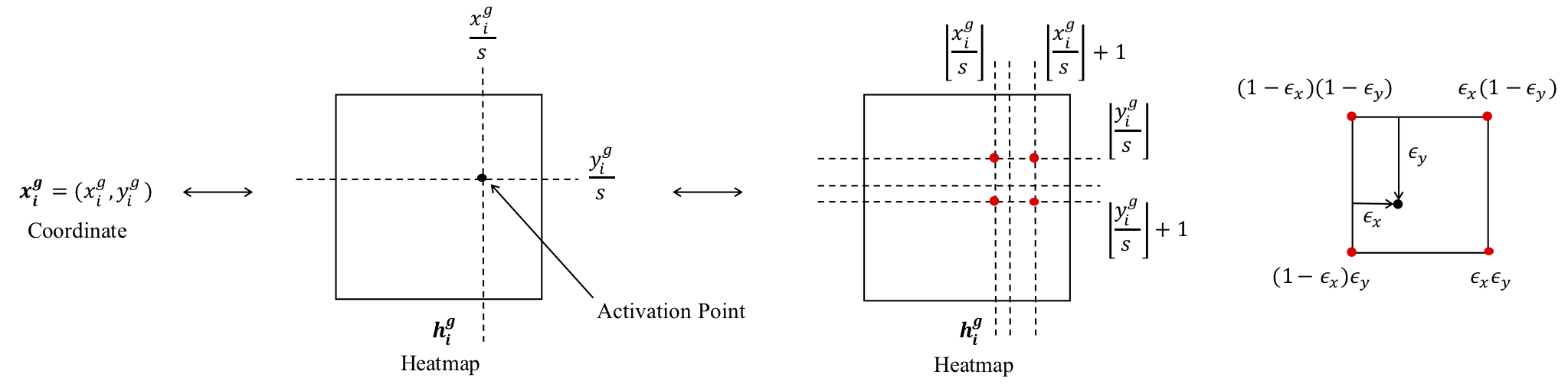}}
\caption{An intuitive example of the encode operation via randomized rounding. When the downsampling stride $s>1$, the ground truth activation point $(x_i^g/s, y_i^g/s)$ usually does not correspond to a single pixel in the heatmap. Therefore, we introduce the randomized rounding operation to assign the ground truth activation point to a set of alternative activation points and the activation probability depends on the fractional part of the ground truth numerical coordinates.}
\label{fig:randomized-rounding}
\end{center}
\end{figure*}

\subsection{Quantization Error}
In this subsection, we first correct the bias in a vanilla quantization system to form an unbiased vanilla quantization system. With the unbiased quantization system, we then provide a tight upper bound on the quantization error for vanilla heatmap regression.

Let $\epsilon_x$ denote the fractional part of $x_i^g/s$, and $\epsilon_y$ denote the fractional part of $y_i^g/s$. Given the downsampling stride of the heatmap $s>1$, we then have
\begin{equation}
\begin{split}
\epsilon_x &=  x_i^g / s - \lfloor x_i^g / s \rfloor, \\ 
\epsilon_y &=  y_i^g / s - \lfloor y_i^g / s \rfloor. 
\end{split}
\end{equation}
Given the assumption of a ``perfect" heatmap prediction model or no heatmap error, i.e., $\boldsymbol{h}_i^p(\boldsymbol{x}) = \boldsymbol{h}_i^g(\boldsymbol{x})$, we then have the predicted numerical coordinates 
\begin{equation}
x_i^p/s = 
\begin{cases}
\lfloor x_i^g/s \rfloor & \quad \text{if } \epsilon_x < t, \\
\lfloor x_i^g/s \rfloor + 1 & \quad \text{otherwise}, \nonumber
\end{cases}
\end{equation}
\begin{equation}
y_i^p/s = 
\begin{cases}
\lfloor y_i^g/s \rfloor & \quad \text{if } \epsilon_y < t, \\
\lfloor y_i^g/s \rfloor + 1 & \quad \text{otherwise}. \nonumber
\end{cases}
\end{equation}
If data samples satisfy the i.i.d. assumption and the fractional parts $\epsilon_x, \epsilon_y \in \mathbb{U}(0,1)$, the bias of $\boldsymbol{x}_i^p$ as an estimator of $\boldsymbol{x}_i^g$ can then be evaluated as 

\begin{equation}
\begin{split}
\mathbb{E}\left(x_i^p/s - x_i^g/s\right) &= \mathbb{E}\left(\mathbf{1}{\{\epsilon_x<t\}}(-\epsilon_x) + \mathbf{1}{\{\epsilon_x\geq t\}}(1-\epsilon_x)\right) \\ \nonumber
 &= 0.5 - t.
 \end{split}
\end{equation}
Considering that $x_i^g, y_i^g$ are independent variables, we thus have the quantization bias in the vanilla quantization system as follows:

\begin{equation}
\label{eq:bias}
\begin{split}
    \mathbb{E}\left(\boldsymbol{x}_i^p/s - \boldsymbol{x}_i^g/s \right) &=  \left(\mathbb{E}\left(x_i^p/s-x_i^g/s\right),~\mathbb{E}\left(y_i^p/s-y_i^g/s\right)\right) \\\nonumber
    &= \left(0.5 - t,~0.5 - t\right).
\end{split}
\end{equation}
Therefore, only the encode operation in~\eqref{eq:quantization}, i.e., the \textbf{round} operation, is unbiased. Furthermore, given $\forall t~\in~[0,1]$ for the encode operation in~\eqref{eq:quantization}, we can correct the bias of the encode operation with a shift on the decode operation, i.e.,
\begin{equation}
    \label{eq:argmax:unbiased}
    \boldsymbol{x}_i^p \in s * \left(\underset{\boldsymbol{x}}{\arg\max}~\left\{\boldsymbol{h}_i^p(\boldsymbol{x})\right\} + t-0.5\right).
\end{equation}

\noindent For simplicity, we use the \textbf{round} operation, i.e., $t=0.5$, to form an unbiased quantization system as our baseline. Though the vanilla quantization system defined by~\eqref{eq:quantization} and~\eqref{eq:argmax:unbiased} is unbiased, it causes non-invertible localization error. An intuitive explanation for this is that the encode operation in ~\eqref{eq:quantization} directly discards the fractional part of the ground truth numerical coordinates, thus making it impossible for the decode operation to accurately reconstruct the numerical coordinates.
\begin{theorem}
\label{thm:quantization}
Given an unbiased quantization system defined by the encode operation in~\eqref{eq:quantization} and the decode operation in~\eqref{eq:argmax:unbiased}, we then have that the quantization error tightly upper bounded, i.e.,
\begin{equation}
    \| \boldsymbol{x}_i^p - \boldsymbol{x}_i^g \|_2 \leq \sqrt{2}s/2, \nonumber
\end{equation}
where $s>1$ indicates the downsampling stride of the heatmap.
\begin{proof}
In Appendix.
\end{proof}
\end{theorem}

\noindent From Theorem~\ref{thm:quantization}, we know that the vanilla quantization system defined by~\eqref{eq:quantization} and~\eqref{eq:argmax:unbiased} will cause non-invertible quantization error and that the upper bound of the quantization error linearly depends on the downsampling stride of the heatmap. As a result, given the heatmap regression model, it will cause extremely large localization error for large faces in the original input image, making it a significant problem in many important face-related applications such as face makeup, face swapping, and face reenactment.

\begin{figure*}[!ht]
\begin{center}
\centerline{\includegraphics[width=\linewidth]{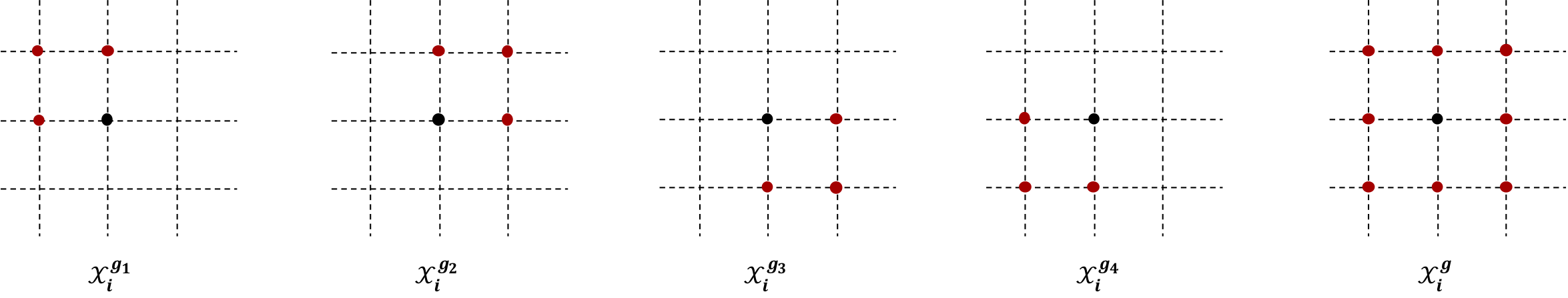}}
\caption{An example of different possible sets of alternative activation points.}
\label{fig:decode:cases}
\end{center}
\end{figure*}

\subsection{Randomized Rounding}
\label{sec:method:randomized-rounding}

In vanilla heatmap regression, each numerical coordinate corresponds to a \textbf{single activation point} in the heatmap, while the indices of the activation point are all integers. As a result, the fractional part of the numerical coordinate is usually ignored during the encode process, making it an inherent drawback of heatmap regression for sub-pixel localization. To retain the fractional information when using heatmap representations, we utilize multiple activation points around the ground truth activation point. Inspired by the randomized rounding method~\cite{raghavan1987randomized}, we address the quantization error in vanilla heatmap regression by using a probabilistic approach. Specifically, we encode the fractional part of the numerical coordinate to different activation points with different activation probabilities. An intuitive example is shown in Fig.~\ref{fig:randomized-rounding}. 

We describe the proposed quantization system as follows. Given the ground truth numerical coordinate $\boldsymbol{x}_i^g = (x_i^g, y_i^g)$ and a downsampling stride of the heatmap $s>1$, the ground truth activation point in the heatmap is $(x_i^g/s, y_i^g/s)$, which are usually floating-point numbers, and we are unable to find the corresponding pixel in the heatmap. If we ignore the fractional part $(\epsilon_x, \epsilon_y)$ using a typical integer quantization operation, e.g., \textbf{round}, the ground truth activation point will be approximated by one of the activation points around the ground truth activation point, i.e., $\left(\lfloor x_i^g/s \rfloor, \lfloor y_i^g/s \rfloor\right)$, $\left(\lfloor x_i^g/s \rfloor+1, \lfloor y_i^g/s \rfloor\right)$, $\left(\lfloor x_i^g/s \rfloor, \lfloor y_i^g/s \rfloor+1\right)$, and $\left(\lfloor x_i^g/s \rfloor+1, \lfloor y_i^g/s \rfloor+1\right)$. However, the above process is not invertible. To address this, we randomly assign the ground truth activation point to one of the \textbf{alternative activation points} around the ground truth activation point, and the activation probability is determined by the fractional part of the ground truth activation point as follows:
\begin{equation}
\label{eq:quantization:prob}
\begin{split}
P\left\{\boldsymbol{h}_i^g\left(\lfloor x_i^g/s \rfloor, \lfloor y_i^g/s \rfloor\right)=1\right\} &= (1-\epsilon_x)(1-\epsilon_y), \\
P\left\{\boldsymbol{h}_i^g\left(\lfloor x_i^g/s \rfloor+1, \lfloor y_i^g/s \rfloor\right)=1\right\} &= \epsilon_x(1-\epsilon_y), \\
P\left\{\boldsymbol{h}_i^g\left(\lfloor x_i^g/s \rfloor, \lfloor y_i^g/s \rfloor+1\right)=1\right\} &= (1-\epsilon_x)\epsilon_y, \\
P\left\{\boldsymbol{h}_i^g\left(\lfloor x_i^g/s \rfloor+1, \lfloor y_i^g/s \rfloor+1\right)=1\right\} &= \epsilon_x\epsilon_y. 
\end{split}
\end{equation}
To achieve the encode scheme in~\eqref{eq:quantization:prob} in conjunction with current minibatch stochastic gradient descent training algorithms for deep learning models, we introduce a new integer quantization operation via randomized rounding, i.e., \textbf{random-round}:
\begin{equation}
\label{eq:quantization:random}
\boldsymbol{q}(x,s) = 
\begin{cases}
\lfloor x/s \rfloor & \quad \text{if } \epsilon < t, ~t \sim \mathbb{U}(0,1), \\
\lfloor x/s \rfloor + 1 & \quad \text{otherwise}.
\end{cases}
\end{equation}
Given the encode operation in~\eqref{eq:quantization:random}, if we do not consider the heatmap error, we then have the activation probability at $\boldsymbol{x}$:
\begin{equation}
\boldsymbol{h}_i^p(\boldsymbol{x}) = P\left\{\boldsymbol{h}_i^g(\boldsymbol{x})=1\right\}.
\end{equation}
As a result, the fractional part of the ground truth numerical coordinate $(\epsilon_x, \epsilon_y)$ can be reconstructed from the predicted heatmap via the activation probabilities of all activation points, i.e.,
\begin{equation}
\label{eq:decode:random}
\boldsymbol{x}_i^p = s*\left(\sum_{\boldsymbol{x} \in \mathcal{X}_i^g}~\boldsymbol{h}_i^p(\boldsymbol{x})*\boldsymbol{x}\right),
\end{equation}
where $\mathcal{X}_i^g$ indicates the set of activation points around the ground truth activation point, i.e.,
\begin{equation}
\label{eq:decode:4points}
\begin{split}
\mathcal{X}_i^g = \{ &\left(\lfloor x_i^g/s \rfloor, \lfloor y_i^g/s \rfloor\right), \left(\lfloor x_i^g/s \rfloor+1, \lfloor y_i^g/s \rfloor\right), \\
	& \left(\lfloor x_i^g/s \rfloor, \lfloor y_i^g/s \rfloor+1\right), \left(\lfloor x_i^g/s \rfloor+1, \lfloor y_i^g/s \rfloor+1\right)\}.
\end{split}
\end{equation}
\begin{theorem}
\label{thm:quantization:random}
Given the encode operation in~\eqref{eq:quantization:random} and the decode operation in~\eqref{eq:decode:random}, we then have that the 1) encode operation is unbiased; and 2) quantization system is lossless, i.e., there is no quantization error.
\begin{proof}
In Appendix.
\end{proof}
\end{theorem}

\noindent From Theorem~\ref{thm:quantization:random}, we know that the quantization system defined by the encode operation in~\eqref{eq:quantization:random} and the decode operation in~\eqref{eq:decode:random} is unbiased and lossless.

\subsection{Activation Points Selection}
\label{sec:method:activation-points}

The fractional information of the numerical coordinate $(\epsilon_x, \epsilon_y)$ is well-captured by the randomized rounding operation, allowing us to reconstruct the ground truth numerical coordinate $\boldsymbol{x}_i^g$ without the quantization error. However, during the inference phase, the ground truth numerical coordinate $\boldsymbol{x}_i^g$ is unavailable and heatmap error always exists in practice, making it difficult to identify the proper set of ground truth activation points $\mathcal{X}_i^g$. In this section, we describe a way to form a set of alternative activation points in practice.

We introduce two activation point selection methods as follows. The first solution is to estimate all activation points via the points around the maximum activation point. As shown in Fig.~\ref{fig:decode:cases}, given the maximum activation point, we then have four different sets of alternative activation points, $\mathcal{X}_i^{g_1}, \mathcal{X}_i^{g_2}, \mathcal{X}_i^{g_3}, \text{and}~\mathcal{X}_i^{g_4}$. Therefore, given the predicted heatmap in practice, we then take a risk of choosing an incorrect set of alternative activation points. To find a robust set of alternative activation points, we may use all nine activation points around the maximum activation point, i.e., 
\begin{equation}
\label{eq:decode:9points}
\mathcal{X}_i^{g} = \mathcal{X}_i^{g_1} \cup \mathcal{X}_i^{g_2} \cup \mathcal{X}_i^{g_3} \cup \mathcal{X}_i^{g_4}.
\end{equation}
Another solution of alternative activation points  $\mathcal{X}_i^{g}$ is to generalize the \textbf{argmax} operation with the \textbf{argtopk} operation, i.e., we decode the predicted heatmap $\boldsymbol{h}_i^p$ to obtain the numerical coordinate $\boldsymbol{x}_i^p$ according to the top $k$ largest activation points, 
\begin{equation}
\label{eq:decode:topkpoints}
\mathcal{X}_i^{g} = \underset{\boldsymbol{x}}{\arg\text{topk}}(\boldsymbol{h_i^p(\boldsymbol{x})}).
\end{equation}
If there is no heatmap error, the two alternative activation points solutions presented above, i.e., the alternative activation points in~\eqref{eq:decode:4points} and~\eqref{eq:decode:9points}, are equal to each other when using the decode operation in \eqref{eq:decode:random}. Specifically, we find that the activation points in~\eqref{eq:decode:9points} achieve comparable performance to the activation points in~\eqref{eq:decode:topkpoints} when $k=9$. For simplicity, unless otherwise mentioned, we use the set of alternative activation points defined by~\eqref{eq:decode:topkpoints} in this paper. Furthermore, when we take the heatmap error into consideration, the values of different $k$ then forms a trade-off on the selection of activation points, i.e., a larger $k$ will be robust to activation point selection whilst also increasing the risk of noise from the heatmap error. See more discussion in Section~\ref{sec:method:discussion} and the experimental results in Section~\ref{sec:exp:ablation}.

\subsection{Discussion}
\label{sec:method:discussion}

In this subsection, we provide some insights into the proposed quantization system with respect to: 1) the influence of human annotators on the proposed quantization system in practice; and 2) the underlying explanation behind the widely used empirical compensation method ``shift a quarter to the second maximum activation point".

\begin{figure}[!ht]
\begin{center}
\centerline{\includegraphics[width=0.9\linewidth]{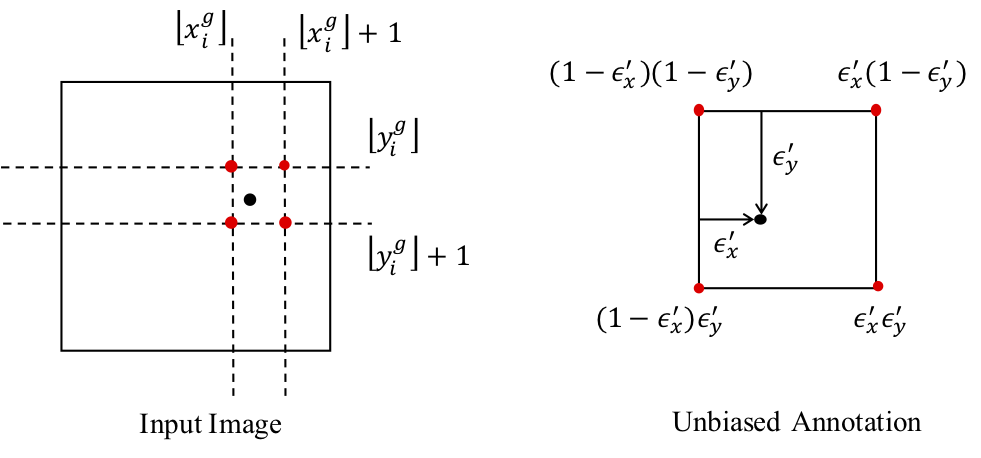}}
\caption{An example of unbiased human annotation for semantic landmark localization.}
\label{fig:annotation}
\end{center}
\end{figure}

\textbf{Unbiased Annotation}. We assume the ``ground truth numerical coordinates'' are always accurate, while the ground truth numerical coordinates are usually labelled by human annotators at the risk of the annotation bias. Given an input image, the ground truth numerical coordinates $\boldsymbol{x}_i^g$ can be obtained by clicking a specific pixel in the image, which is a simple but effective annotation pipeline provided by most image annotation tools. For sub-pixel numerical coordinates, especially in low-resolution input images, the annotators may click either one of all possible pixels around the ground truth numerical coordinates due to human visual uncertainty. As shown in Fig.~\ref{fig:annotation}, clicking any one of the four possible pixels causes annotation error, which corresponds to the fractional part of the ground truth numerical coordinate $(\epsilon_x', \epsilon_y') = \left(x_i^g - \lfloor x_i^g \rfloor, ~y_i^g - \lfloor y_i^g \rfloor\right)$.
Given enough data samples, if the annotators click the pixel according to the following distribution, i.e., 
\begin{equation}
\label{eq:annotation:prob}
\begin{split}
P\left\{\left(\lfloor x_i^g \rfloor, \lfloor y_i^g \rfloor\right)\right\} &= (1-\epsilon_x')(1-\epsilon_y'), \\
P\left\{\left(\lfloor x_i^g \rfloor+1, \lfloor y_i^g \rfloor\right)\right\} &= \epsilon_x'(1-\epsilon_y'), \\
P\left\{\left(\lfloor x_i^g \rfloor, \lfloor y_i^g \rfloor+1\right)\right\} &= (1-\epsilon_x')\epsilon_y', \\
P\left\{\left(\lfloor x_i^g \rfloor+1, \lfloor y_i^g \rfloor+1\right)\right\} &= \epsilon_x'\epsilon_y',
\end{split}
\end{equation}
the fractional part then can be well captured by the heatmap regression model and we refer to it as an unbiased annotation. 
If we take the downsampling stride into consideration, $(\epsilon_x, \epsilon_y)$ is then a joint result of both the downsampling of the heatmap and the annotation process, i.e.,
\begin{equation}
\label{eq:annotation-bias}
(\epsilon_x,~\epsilon_y)~\propto~\left(\epsilon_{x}'/s, ~\epsilon_{y}'/s\right) + (s-1).
\end{equation}
On the one hand, if the heatmap regression model uses a low input resolution (or a large downsampling stride $s\gg1$), the fractional part $(\epsilon_x, \epsilon_y)$ then mainly comes from the downsampling of the heatmap; on the other hand, if the heatmap regression model uses a high input resolution, the annotation process will also have a significant influence on the heatmap regression. Therefore, when using a high input resolution model in practice, a diverse set of human annotators help reduce the bias in annotation process.

\textbf{Empirical Compensation}. ``Shift a quarter to the second maximum activation point" has become an  effective and widely used empirical compensation method for heatmap regression~\cite{newell2016stacked,chen2018cascaded,sun2019deep}, but it still lacks a proper explanation. We thus provide an intuitive explanation according to the proposed quantization system. The proposed quantization system encodes the ground truth numerical coordinates into multiple activation points, and the activation probability of each activation point is decided by the fractional part, i.e., the activation probability indicates the distance between the activation point and the ground truth activation point. Therefore, the ground truth activation point is closer to the $i$-th maximum activation point than the $(i+1)$-th maximum activation point. We demonstrate the averaged activation probabilities for the top $k$ activation points on the WFLW dataset in Table~\ref{table:activation:prob}. 
\begin{table}[ht]
\caption{The activation probabilities of the top $k$ activation points.}
\begin{center}
\begin{tabular}{|>{\centering\arraybackslash}m{0.1\textwidth}|>{\centering\arraybackslash}m{0.05\textwidth}|>{\centering\arraybackslash}m{0.05\textwidth}|>{\centering\arraybackslash}m{0.05\textwidth}|>{\centering\arraybackslash}m{0.05\textwidth}|}
\hline
  & k=1 & k=2 & k=3 & k=4 \\ 
 \hline
  $\boldsymbol{h}_i^p(\boldsymbol{x})$  & 0.44 & 0.26 & 0.17 & 0.13 \\ 
 \hline
  NME(\%) & 6.45 & 5.07 &  4.71 & 4.68 \\
 \hline
\end{tabular}
\end{center}
\label{table:activation:prob}
\end{table}
We find that the marginal improvement decreases as the number of activation points increases, i.e., the second maximum activation points provides the maximum improvement to the reconstruction of the fractional part. This observation partially explains the effectiveness of the compensation method ``shift a quarter to the second maximum activation point", which can be seen as a special case of the proposed method~\eqref{eq:decode:topkpoints} with $k=2$.

Furthermore, the proposed quantization system shares the same motivation with the bilinear interpolation. Specifically, the bilinear interpolation usually aims to find the value of the unknown function $f(x, y)$ given its neighbors $f(x_1,y_1)$, $f(x_1,y_2)$, $f(x_2,y_1)$, and $f(x_2,y_2)$. For the proposed quantization system, we have $f(x,y) = (x,y)$, which indicates the location of landmarks. Specifically, if there is no heatmap error, we then have $x_1 = \lfloor x_i^g/s \rfloor$, $x_2 = \lfloor x_i^g/s \rfloor + 1$, $y_1 = \lfloor y_i^g/s \rfloor$, and  $y_2 = \lfloor y_i^g/s \rfloor + 1$. If we take heatmap error into consideration, the ground truth activation points are usually unknown. Therefore, the number of alternative activation points also controls the trade-off between the robustness of the quantization system and the risk of noise from the heatmap error (Also see details in Section \ref{sec:method:activation-points}).

\begin{table*}[!ht]
\caption{Comparison with State-of-the-Arts on WFLW dataset.}
\label{table:soa:wflw}
\begin{center}
\begin{tabular}{|c|c|c|c|c|c|c|c|}
\hline
\multirow{2}{*}{Method} & \multicolumn{7}{c|}{NME (\%), Inter-ocular} \\
\cline{2-8} &  \textbf{test} & \textbf{pose} &  \textbf{expression} &  \textbf{illumination} &  \textbf{make-up} & \textbf{occlusion} & \textbf{blur} \\
\hline
ESR~\cite{cao2014face}   & 11.13  & 25.88  & 11.47  & 10.49  & 11.05  & 13.75  & 12.20  \\ 
\hline
SDM~\cite{xiong2013supervised}   & 10.29  & 24.10  & 11.45  & 9.32   & 9.38   & 13.03  & 11.28  \\
\hline
CFSS~\cite{zhu2015face}  & 9.07   & 21.36  & 10.09  & 8.30   & 8.74   & 11.76  & 9.96   \\
\hline
DVLN~\cite{wu2017leveraging}  & 6.08   & 11.54  & 6.78   & 5.73   & 5.98   & 7.33   & 6.88   \\
\hline
LAB~\cite{wu2018look} & 5.27  &10.24  & 5.51   & 5.23   & 5.15   & 6.79   & 6.32 \\
\hline
Wing~\cite{feng2018wing} & 5.11 & 8.75   & 5.36   & 4.93   & 5.41   & 6.37   & 5.81 \\
\hline
3DDE~\cite{valle2019face} & 4.68 & 8.62 & 5.21 & 4.65 & 4.60 & 5.77 & 5.41 \\
\hline 
DeCaFA~\cite{Dapogny_2019_ICCV} & 4.62 & 8.11 & 4.65 & 4.41 & 4.63 & 5.74 & 5.38 \\
\hline
HRNet~\cite{sun2019high} & 4.60 & 7.86	& 4.78 & 4.57 &  4.26 & 5.42 &	5.36 \\
\hline
AVS~\cite{qian2019aggregation} & 4.39 & 8.42 & 4.68 & 4.24 & 4.37 & 5.60 & 4.86 \\
\hline
LUVLi~\cite{kumar2020luvli} & 4.37 & - & - & - & - & - & - \\
\hline
AWing~\cite{wang2019adaptive} & 4.21 & 7.21 & 4.46 & 4.23 & 4.02 & 4.99 & 4.82 \\
\hline
H3R~(\textbf{ours}) & \textbf{3.81} & \textbf{6.45} & \textbf{4.07} & \textbf{3.70} & \textbf{3.66} &  \textbf{4.48} & \textbf{4.30} \\
\hline
\end{tabular}
\end{center}
\end{table*}

\begin{table*}[!ht]
\caption{Comparison with State-of-the-Arts on 300W dataset.}
\label{table:soa:300w}
\begin{center}
\begin{tabular}{|c|c|c|c|c|c|c|c|c|c|c|}
\hline
\multirow{2}{*}{Method} & \multicolumn{4}{c|}{NME (\%), Inter-ocular} & \multicolumn{4}{c|}{NME (\%), Inter-pupil}  \\
\cline{2-9} & \textbf{private} & \textbf{full} & \textbf{common} & \textbf{challenge} & \textbf{private}  & \textbf{full} & \textbf{common} & \textbf{challenge} \\
\hline 
SAN~\cite{dong2018style} & - &  3.98 & 3.34 & 6.60 & - & - & - & -  \\
 \hline 
DAN~\cite{kowalski2017deep} & 4.30 & 3.59 & 3.19 & 5.24 & - & 5.03 & 4.42 & 7.57  \\
\hline 
SHN~\cite{yang2017stacked} & 4.05 & - & - & - & - & 4.68 & 4.12 & 7.00 \\
\hline
LAB~\cite{wu2018look} & - & 3.49 & 2.98 & 5.19 & - & 4.12 & 3.42 & 6.98  \\
\hline
Wing~\cite{feng2018wing} & - & - & - & -  & - & \textbf{4.04} & \textbf{3.27} & 7.18 \\
\hline
DeCaFA~\cite{Dapogny_2019_ICCV} & - & 3.39 & 2.93 & 5.26 & - & - & - & -  \\
\hline
DFCE~\cite{valle2018deeply} & 3.88 & 3.24 & 2.76 & 5.22 &- & 4.55 & 3.83 & 7.54 \\
\hline
 AVS~\cite{qian2019aggregation} & - & 3.86 & 3.21 & 6.49 & - & 4.54 & 3.98 & 7.21  \\
 \hline
HRNet~\cite{sun2019high} & 3.85 &  3.32 & 2.87 & 5.15 & - & - & - & - \\
\hline
HG-HSLE~\cite{Zou_2019_ICCV} & - & 3.28 & 2.85 & 5.03 & - & 4.59 & 3.94 & 7.24 \\
\hline
LUVLi~\cite{kumar2020luvli} & - & 3.23 & 2.76 & 5.16 & - & - & - & - \\
\hline
3DDE~\cite{valle2019face} & 3.73 & 3.13 & 2.69 & 4.92 & - & 4.39 & 3.73 & 7.10 \\
\hline
AWing~\cite{wang2019adaptive} & 3.56 & 3.07 & 2.72 & \textbf{4.52} & - & 4.31 & 3.77 & \textbf{6.52} \\
\hline 
H3R (\textbf{ours}) & \textbf{3.48} & \textbf{3.02}  & \textbf{2.65}  & 4.58 & \textbf{5.07} & 4.24 & 3.67 & 6.60 \\
\hline
\end{tabular}
\end{center}
\end{table*}

\section{Facial Landmark Detection}

In this section, we perform facial landmark detection experiments. We first introduce widely used facial landmark detection datasets. We then describe the implementation details of our proposed method. Finally, we present our experimental results on different datasets and perform comprehensive ablation studies on the most challenging dataset.

\subsection{Datasets}
We use four widely used facial landmark detection datasets:
\begin{itemize}
    \item \textbf{WFLW}~\cite{wu2018look}. WFLW contains $10,000$ face images, including $7,500$ training images and $2,500$ testing images, with $98$ manually annotated facial landmarks. All face images are selected from the WIDER Face dataset~\cite{yang2016wider}, which contains face images with large variations in scale, expression, pose, and occlusion. 
    \item \textbf{300W}~\cite{sagonas2013300}. 300W contains $3,148$ training images, including $337$ images from AFW~\cite{zhu2012face}, $2,000$ images from the training set of HELEN~\cite{le2012interactive}, and $811$ images from the training set of LFPW~\cite{belhumeur2013localizing}. For testing, there are four different settings: 1) \textbf{common}: $554$ images, including $330$ and $224$ images from the testsets of HELEN and LFPW, respectively; 2) \textbf{challenge}: $135$ images from IBUG; 3) \textbf{full}: $689$ images as a combination of \textbf{common} and \textbf{challenge}; and 4) \textbf{private}: $600$ indoor/outdoor images. All images are manually annotated with $68$ facial landmarks.
    \item \textbf{COFW}~\cite{burgos2013robust}. COFW contains $1,852$ images, including $1,345$ training and $507$ testing images. All images are manually annotated with $29$ facial landmarks.
    \item \textbf{AFLW}~\cite{koestinger2011annotated}. AFLW contains $24,386$ face images, including $20,000$ images for training and $4,836$ images for testing. For testing, there are two settings: 1) \textbf{full}: all $4,836$ images for testing; and 2) \textbf{front}: $1,314$ frontal images selected from the \textbf{full} set. All images are manually annotated with $21$ facial landmarks. For fair comparison, we use $19$ facial landmarks, i.e., the landmarks on two ears are ignored.
\end{itemize}

\subsection{Evaluation Metrics}

We use the normalized mean error (NME) as the evaluation metric in this paper, i.e.,
\begin{equation}
    \text{NME} = \mathbb{E}\left(\frac{\|\boldsymbol{x}_i^p-\boldsymbol{x}_i^g\|_2}{d}\right),
\end{equation}
where $d$ indicates the normalization distance. For fair comparison, we report the performances on WFLW, 300W, and COFW using two the normalization methods,~\textbf{inter-pupil distance} (the distance between the eye centers) and~\textbf{inter-ocular distance} (the distance between the outer eye corners). We report the performance on AFLW using the size of the face bounding box as the normalization distance, i.e., the normalization distance can be evaluated by $d = \sqrt{w*h}$, where $w$ and $h$ indicate the width and height of the face bounding box, respectively.

\subsection{Implementation Details}

We implement the proposed heatmap regression method for facial landmark detection using PyTorch~\cite{NIPS2019_9015}. Following the practice in~\cite{sun2019high}, we use HRNet~\cite{sun2019deep} as our backbone network, which is an efficient counterpart of ResNet~\cite{he2016deep}, U-Net~\cite{ronneberger2015u}, and Hourglass~\cite{newell2016stacked} for semantic landmark localization. Unless otherwise mentioned, we use HRNet-W18 as the backbone network in our experiments. All face images are cropped and resized to $256\times256$ pixels and the downsampling stride of the feature map is $4$ pixels. For training, we perform widely-used data augmentation for facial landmark detection as follows We horizontally flip all training images with probability $0.5$ and randomly change the brightness ($\pm0.125$), contrast ($\pm0.5$), and saturation ($\pm0.5$) of each image. We then randomly rotate the image ($\pm30^{\circ}$), rescale the image ($\pm0.25$), and translate the image ($\pm16$ pixels). We also randomly erase a rectangular region in the training image~\cite{zhong2017random}. All our models are initialized from the weights pretrained on ImageNet~\cite{deng2009imagenet}. We use the Adam optimizer~\cite{kingma2015adam} with batch size $16$. The learning rate starts from $0.001$ and is divided by $10$ for every $60$ epochs, with $150$ training epochs in total. During the testing phase, we horizontally flip testing images as the data augmentation and average the predictions.

\begin{table}[!ht]
\caption{Comparison with State-of-the-Arts on COFW dataset.}
\label{table:soa:cofw}
\begin{center}
\begin{tabular}{|>{\centering\arraybackslash}m{0.1\textwidth}|>{\centering\arraybackslash}m{0.1\textwidth}|>{\centering\arraybackslash}m{0.1\textwidth}|}
\hline
\multirow{2}{*}{Method} &  \multicolumn{2}{c|}{NME (\%)} \\
\cline{2-3} & Inter-ocular & Inter-pupil \\
\hline 
SHN~\cite{yang2017stacked} & - & 5.60 \\
\hline
LAB~\cite{wu2018look} & - & 5.58  \\
\hline
DFCE~\cite{valle2018deeply} & - & 5.27 \\
\hline
3DDE~\cite{valle2019face} & - & 5.11  \\
\hline
Wing~\cite{feng2018wing} & - & 5.07  \\
\hline
HRNet~\cite{sun2019high} & 3.45 & -\\
\hline
AWing~\cite{wang2019adaptive} & - & 4.94 \\
\hline
H3R (\textbf{ours}) & \textbf{3.15} & \textbf{4.55} \\
\hline
\end{tabular}
\end{center}
\end{table}

\begin{table}[!ht]
\caption{Comparison with State-of-the-Arts on AFLW dataset.}
\label{table:soa:aflw}
\begin{center}
\begin{tabular}{|>{\centering\arraybackslash}m{0.1\textwidth}|>{\centering\arraybackslash}m{0.1\textwidth}|>{\centering\arraybackslash}m{0.1\textwidth}|}
\hline
\multirow{2}{*}{Method} &  \multicolumn{2}{c|}{NME (\%)}  \\
\cline{2-3} & \textbf{full} & \textbf{front} \\
\hline
DFCE~\cite{valle2018deeply} & 2.12 & - \\
\hline
3DDE~\cite{valle2019face} & 2.01 & - \\
\hline
SAN~\cite{dong2018style} & 1.91 & 1.85 \\
\hline
LAB~\cite{wu2018look} & 1.85 &  1.62 \\
\hline
Wing~\cite{feng2018wing} & 1.65 & -  \\
\hline
HRNet~\cite{sun2019high} & 1.57 & 1.46 \\
\hline
LUVLi~\cite{kumar2020luvli} & 1.39 & 1.19\\
\hline 
H3R (\textbf{ours}) & \textbf{1.27} & \textbf{1.11} \\
\hline
\end{tabular}
\end{center}
\end{table}

\subsection{Comparison with Current State-of-the-Art}
\label{sec:exp:soa}

To demonstrate the effectiveness of the proposed method, we compare it with recent state-of-the-art facial landmark detection methods as follows. As shown in Table~\ref{table:soa:wflw}, the proposed method outperforms recent state-of-the-art methods on the most challenging dataset, WFLW, with a clear margin for all different settings. For the 300W dataset, we try to report the performances under different settings for fair comparison. As shown in Table~\ref{table:soa:300w}, the proposed method achieves comparable performances for all different settings. Specifically, LAB~\cite{wu2018look} uses the boundary information as the auxiliary supervision; compared to Wing~\cite{feng2018wing}, which uses the coordinate regression for semantic landmark localization, the heatmap-based methods usually achieve better performance on the challenge subset. In Table~\ref{table:soa:cofw}, we see that the proposed method outperforms recent state-of-the-art methods with a clear margin on COFW dataset. AFLW captures a wide range of different face poses, including both frontal faces and non-frontal faces. As shown in Table~\ref{table:soa:aflw}, the proposed method achieves consistent improvements for both frontal faces and non-frontal faces, suggesting robustness across different face poses.

\subsection{Ablation Studies}
\label{sec:exp:ablation}

To better understand the proposed quantization system in different settings, we perform ablation studies the most challenging dataset, WFLW~\cite{wu2018look}. 
\begin{figure}[!ht]
\begin{center}
\centerline{\includegraphics[width=\linewidth]{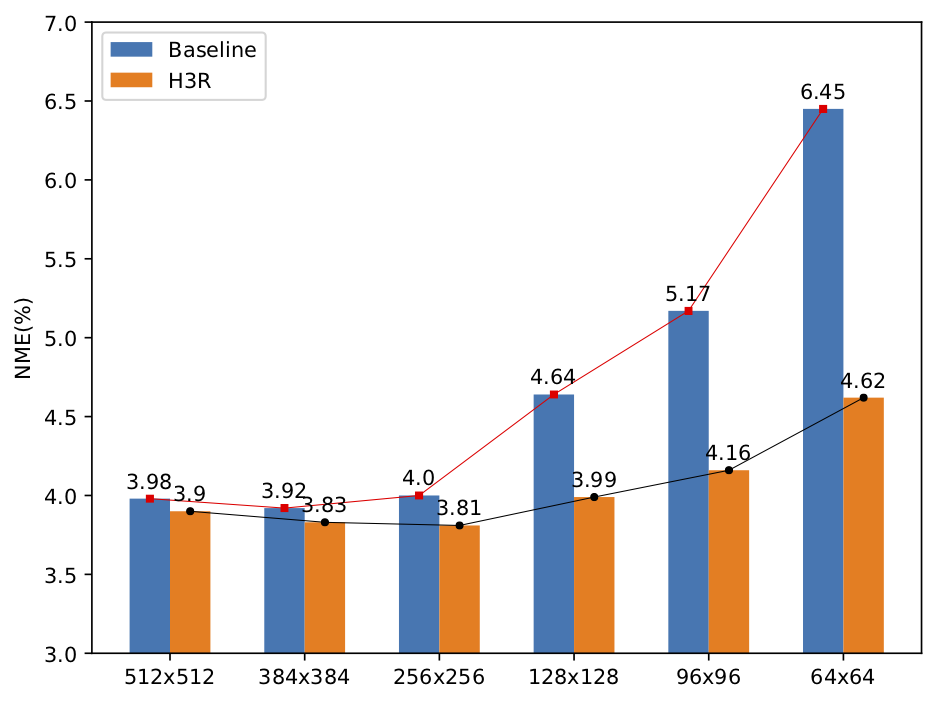}}
\end{center}
\caption{The influence of different input resolutions.}
\label{fig:exp:ablation:resolution}
\end{figure}

\textbf{The influence of different input resolutions}. Heatmap regression models use a fixed input resolution, e.g., $256\times256$ pixels, but training and testing images usually represent a wide range of resolutions, e.g., most faces in the WFLW dataset have an inter-ocular distance of between $30$ and $120$ pixels. Therefore, we compare the proposed method with the baseline using an input resolution from $64\times64$ pixels to $512\times512$ pixels, i.e., a heatmap resolution from $16\times16$ pixels to $128\times128$ pixels. In Fig.~\ref{fig:exp:ablation:resolution}, the proposed method significantly improves heatmap regression performance when using a low input resolution. The increasing number of high-resolution images/videos in real-world applications is a challenge with respect to the computational cost and device memory needed to overcome the problem of sub-pixel localization by increasing the input resolution of deep learning-based heatmap regression models. For example, in the film industry, it has sometimes become necessary to swap the appearance of a target actor and a source actor to generate higher fidelity video frames in visual effects, especially when an actor is unavailable for some scenes~\cite{jacek2020high}. The manipulation of actor faces in video frames relies on accurate localization of different facial landmarks and is performed at megapixel resolution, inducing a huge computational cost for extensive frame-by-frame animation. Therefore, instead of using high-resolution input images, our proposed method delivers another efficient solution for dealing with accurate semantic landmark localization.

\begin{table*}[!ht]
\caption{The influence of different numbers of alternative activation points when using binary heatmap.}
\label{table:exp:ablation:topk}
\begin{center}
\begin{tabular}{|c|c|c|c|c|c|c|c|c|c|c|c|c|c|c|c|}
\hline
\multirow{2}{*}{Resolution} & \multicolumn{13}{c|}{NME (\%)} \\
\cline{2-14} & $k=1$ & $k=2$ &  $k=3$  &  $k=4$  & $k=5$ & $k=6$  & $k=7$ & $k=8$ & $k=9$  &  $k=10$ &  $k=11$  &  $k=12$ & \textbf{best} \\
\hline 
$512\times512$ & 3.980 & 3.946  & 3.930 & 3.923 & 3.916 & 3.912 & 3.909 & 3.906 & 3.903 & 3.901 & 3.898 & 3.897 & \textbf{3.890}/$k=25$ \\
\hline
$384\times384$  & 3.932 & 3.875 & 3.855 & 3.850 & 3.842 & 3.837 & 3.833 & 3.830 & 3.828 & 3.827 & 3.826 & 3.826 & \textbf{3.825}/$k=14$ \\
\hline
$256\times256$  &4.005 & 3.881 & 3.836 & 3.832 & 3.819 & 3.815 & 3.810 & 3.808 & \textbf{3.807} & 3.807 & 3.807 & 3.808 & \textbf{3.807}/$k=9$\\
\hline
$128\times128$ & 4.637 & 4.164 & 4.029 & 4.023 & 3.997 & 3.991 & \textbf{3.988} & 3.989 & 3.991 & 3.994 & 3.998 & 4.002 & \textbf{3.988}/$k=7$ \\
\hline
\end{tabular}
\end{center}
\end{table*}

\begin{table*}[!ht]
\caption{The influence of different numbers of alternative activation points when using Gaussian heatmap.}
\label{table:exp:ablation:topk:gaussian}
\begin{center}
\begin{tabular}{|c|c|c|c|c|c|c|c|c|c|c|c|c|c|c|c|}
\hline
\multirow{2}{*}{} & \multicolumn{13}{c|}{NME (\%)} \\
\cline{2-14} & $k=1$ & $k=2$ &  $k=3$  &  $k=4$  & $k=5$ & $k=6$  & $k=7$ & $k=8$ & $k=9$  &  $k=10$ &  $k=11$  &  $k=12$ & \textbf{best} \\
\hline 
$\sigma=0.0$ & 4.235 & 4.102 & 4.069 & \textbf{4.065} & 4.090 & 4.202 & 4.427 & 4.775 & 5.219 & 5.719 & 6.256 & 6.802 & \textbf{4.065}/$k=4$  \\
\hline
$\sigma=0.5$ & 4.162 & 4.045 & 4.010 & 4.008 & 3.991 & 3.990 & \textbf{3.988} & 3.992 & 3.998 & 4.014 & 4.047 & 4.108 & \textbf{3.988}/$k=7$ \\
\hline
$\sigma=1.0$ & 4.037 & 3.928 & 3.898 & 3.908 & 3.871 & 3.877 & 3.876 &  3.876 & 3.871 & 3.861 & \textbf{3.855} & 3.857 & \textbf{3.855}/$k=11$ \\
\hline
$\sigma=1.5$ & 4.032 & 3.927 & 3.901 & 3.918 & 3.873 & 3.879 & 3.885 & 3.882 & 3.878 & 3.864 & 3.857 & 3.860 & \textbf{3.855}/$k=18$ \\
\hline
$\sigma=2.0$ & 4.086 & 3.983 & 3.953 & 3.969 & 3.923 & 3.931 & 3.937 & 3.931 & 3.925 & 3.909 & 3.902 & 3.907 & \textbf{3.894}/$k=25$ \\ 
\hline
\end{tabular}
\end{center}
\end{table*}

\begin{table*}[!ht]
\caption{The influence of different face "bounding box" annotation policies.}
\label{table:exp:ablation:bbox}
\begin{center}
\begin{tabular}{|c|c|c|c|c|c|c|c|c|}
\hline
\multicolumn{2}{|c|}{ Policy} & \multicolumn{7}{c|}{NME (\%), Inter-ocular} \\
\hline
\textbf{training} & \textbf{testing} &  \textbf{test} & \textbf{pose} &  \textbf{expression} &  \textbf{illumination} &  \textbf{make-up} & \textbf{occlusion} & \textbf{blur} \\
\hline
\textbf{P1} & \textbf{P1}  & 3.81 & 6.45 & 4.07 & 3.70 & 3.66 & 4.48 & 4.30 \\
\hline
\textbf{P2} & \textbf{P2}  & 3.95 & 6.74 & 4.17 & 3.89 & 3.81 & 4.73 & 4.51 \\
\hline
\textbf{P1} & \textbf{P2}  & 5.34 & 9.68 & 5.24 & 5.25 & 5.57 & 6.55 & 6.14 \\
\hline
\textbf{P2} & \textbf{P1}  & 4.04 & 6.90 & 4.24 & 3.99 & 3.90 & 4.87 & 4.65 \\
\hline
\end{tabular}
\end{center}
\end{table*}

\textbf{The influence of different numbers of alternative activation points}. In the proposed quantization system, the activation probability indicates the distance between activation point to the ground truth activation point $(x_i^g/s, y_i^g/s)$. If there is no heatmap error, the alternative activation points in \eqref{eq:decode:topkpoints} then give the same result as in \eqref{eq:decode:4points}. If the heatmap error cannot be ignored, there will be a trade-off on the number of alternative activation points: 1) a small $k$ increases the risk of missing the ground truth alternative activation points; 2) a large $k$ introduces the noise from irrelevant activation points, especially for large heatmap error. We demonstrate the performance of the proposed method by using different numbers of alternative activation points in Table~\ref{table:exp:ablation:topk}. Specifically, we see that 1) when using a high input resolution, the best performance is achieved with a relatively large $k$; and 2) the performance is smooth near the optimal number of alternative activation points, making it easy to find a proper $k$ for validation data. 
As introduced in Section~\ref{sec:preliminary}, binary heatmaps can be seen as a special case of Gaussian heatmaps with standard deviation $\sigma=0$. Considering that Gaussian heatmaps have been widely used in semantic landmark localization applications, we generalize the proposed quantization system to Gaussian heatmaps and demonstrate the influence of different numbers of alternative activation points in Table~\ref{table:exp:ablation:topk:gaussian}. Specifically, we see that 1) when applying the proposed quantization system to the model using Gaussian heatmap, it achieves comparable performance to the model using binary heatmap; and 2) the optimal number of alternative activation points increases with the standard deviation $\sigma$.

\begin{figure}[!ht]
\begin{center}
\centerline{\includegraphics[width=\linewidth]{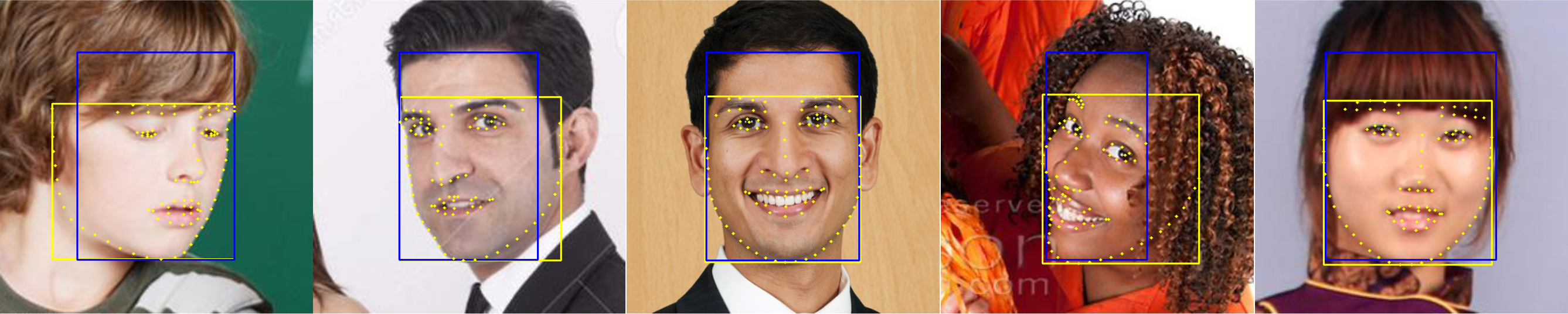}}
\end{center}
\caption{A comparison between different "bounding box" annotation policies. \textbf{P1}: the yellow bounding boxes. \textbf{P2}: the blue bounding boxes.}
\label{fig:exp:ablation:bbox}
\end{figure}

\textbf{The influence of different ``bounding box" annotation policies}. For facial landmark detection, a reference bounding box is required to indicate the position of the facial area. However, there is a performance gap when using different reference bounding boxes~\cite{wang2019adaptive}.  A comparison between two widely used ``bounding box" annotation policies is shown in Fig.~\ref{fig:exp:ablation:bbox}, and we introduce two different bounding box annotation policies as follows:
\begin{itemize}
\item \textbf{P1}: This annotation policy is usually used in semantic landmark localization tasks, especially in  facial landmark localization. Specifically, the rectangular area of the bounding box tightly encloses a set of pre-defined facial landmarks.
\item \textbf{P2}: This annotation policy has been widely used in face detection datasets~\cite{yang2016wider}. The bounding box contains the areas of the forehead, chin, and cheek. For the occluded face, the bounding box is estimated by the human annotator based on the scale of occlusion.
\end{itemize}

\begin{figure*}[!ht]
\begin{center}
\centerline{\includegraphics[width=\linewidth]{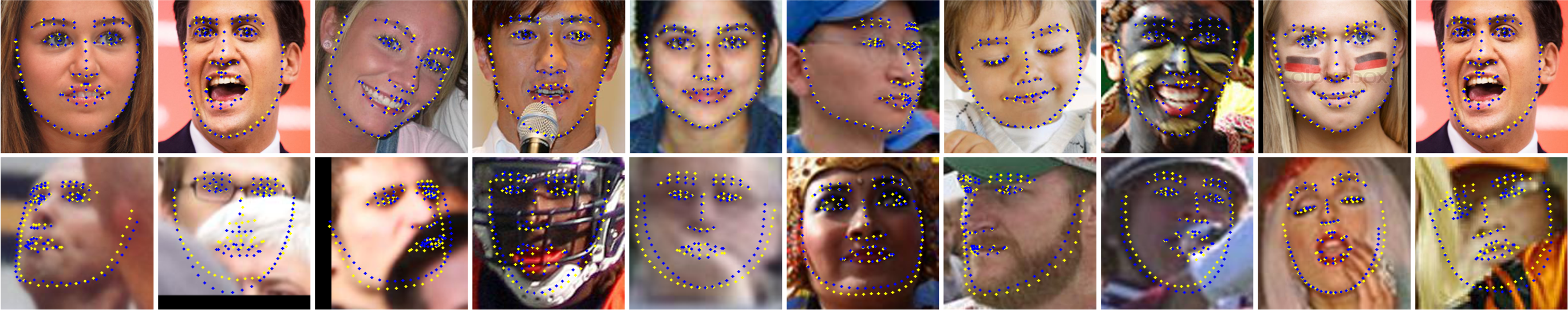}}
\end{center}
\caption{Qualitative results from the test split of the WFLW dataset (best view in color). \textbf{Blue}: the ground truth facial landmarks. \textbf{Yellow}: the predicted facial landmarks. The first row shows some ``good" cases and the second row shows some ``bad" cases.}
\label{fig:demo}
\end{figure*}

\noindent We demonstrate the experimental results using different annotation policies in Table~\ref{table:exp:ablation:bbox}. Specifically, we find that the policy \textbf{P1} usually achieves better results, possibly because the occluded forehead (e.g., hair) introduces additional variations to the face bounding boxes when using the policy \textbf{P2}. Furthermore, the model trained using the policy \textbf{P2} is more robust to different bounding box policy during testing, suggesting its robustness to inaccurate bounding boxes from face detection algorithms.

\textbf{Qualitative Analysis}. As shown in Fig.~\ref{fig:demo}, we present some ``good" and ``bad"  facial landmark detection examples according to NME. Specifically, for the good cases presented in the first row, most images are of high quality; For the bad cases in the second row, most images contain heavy blurring and/or occlusion, making it difficult to accurately identify the contours of different facial parts. 

\begin{table*}[!ht]
\caption{Results on MPII Human Pose dataset. In each block, the first row with ``$-$" indicates the baseline method, i.e., $k=1$; The second row with ``$*$" indicates the compensation method, i.e., ``shift a quarter to the second maximum activation point".}
\label{table:exp:pose:mpii}
\begin{center}
\begin{tabular}{|c|c|c|c|c|c|c|c|c|c|c|c|}
\hline
Backbone & Input & H3R & Head & Shoulder & Elbow & Wrist &  Hip & Knee & Ankle & Mean & Mean@0.1 \\
\hline
ResNet-50 & $256\times256$ & - & 96.3 & 95.2 & 89.0 &  82.9 & 88.2 & 83.7 &  79.4 & 88.4 & 29.8 \\
ResNet-50 & $256\times256$ & * & 96.4 & 95.3 & 89.0 &  83.2 & 88.4 & 84.0 &  79.6 & 88.5 & 34.0 \\
ResNet-50 & $256\times256$ & \checkmark & \textbf{96.3} & \textbf{95.2} & \textbf{88.8} & \textbf{83.4} & \textbf{88.5} & \textbf{84.3} & \textbf{79.8} & \textbf{88.6} & \textbf{34.9} \\
\hline
ResNet-101 & $256\times256$ & - & 96.7 & 95.8 &  89.3 &  84.2 & 87.9 & 84.2 &  80.7 & 88.9 & 30.0  \\
ResNet-101 & $256\times256$ & * & 96.9 & 95.9 &  89.5 &  84.4 & 88.4 & 84.5 &  80.7 & 89.1 & 34.0  \\
ResNet-101 & $256\times256$ & \checkmark & \textbf{96.7} & \textbf{96.0} & \textbf{89.3} & \textbf{84.4} & \textbf{88.5} & \textbf{84.3} & \textbf{80.6} & \textbf{89.1}& \textbf{35.0} \\
\hline 
ResNet-152 & $256\times256$ & - & 97.0 & 95.8 &  89.9 &  84.7 & 89.1 & 85.4 &  81.3 & 89.5 & 31.0 \\
ResNet-152 & $256\times256$ & * & 97.0 & 95.9 &  90.0 &  85.0 & 89.2 & 85.3 &  81.3 & 89.6 & 35.0 \\
ResNet-152 & $256\times256$ &  \checkmark &  \textbf{96.8} &  \textbf{95.9} &  \textbf{90.1} &  \textbf{84.9} &  \textbf{89.3} &  \textbf{85.3} &  \textbf{81.3} &  \textbf{89.6} &  \textbf{36.2}\\
\hline
HRNet-W32 & $256\times256$ & - & 97.0 & 95.7 &  90.1 & 86.3 & 88.4 & 86.8 &  82.9 & 90.1 &  32.8\\
HRNet-W32 & $256\times256$ & * & 97.1 & 95.9 &  90.3 & 86.4 & 89.1 & 87.1 &  83.3 & 90.3 &  37.7\\
HRNet-W32 & $256\times256$ & \checkmark &  \textbf{97.1} &  \textbf{96.1} &  \textbf{90.8} &  \textbf{86.1} &  \textbf{89.2} &  \textbf{86.4} &  \textbf{82.6} &  \textbf{90.3} &  \textbf{39.3} \\
\hline
\end{tabular}
\end{center}
\end{table*}

\begin{table*}[!ht]
\caption{Results on COCO validation set.  In each block, the first row with ``$-$" indicates the baseline, i.e., $k=1$; The second row with ``$*$" indicates the compensation method, i.e., ``shift a quarter to the second maximum activation point", which can be seen as a special case of H3R with $k=2$.}
\label{table:exp:pose:coco}
\begin{center}
\begin{tabular}{|c|c|c|c|c|c|c|c|c|c|c|c|c|}
\hline
Backbone & Input  & H3R & AP & Ap .5 & AP .75 & AP (M) & AP (L) & AR & AR .5 & AR .75 & AR (M) & AR (L) \\
\hline
HRNet-W32 & $192\times128$  & - & 0.674 & 0.890 &  0.771 &  0.648 &  0.732 & 0.739 & 0.932 &  0.828 &  0.700 &  0.795 \\
HRNet-W32 & $192\times128$  & * & 0.710 & 0.892 & 0.792 & 0.682 & 0.771 & 0.770 & 0.933 & 0.844 & 0.732 & 0.827 \\
HRNet-W32 & $192\times128$  & \checkmark& \textbf{0.720} & \textbf{0.892} & \textbf{0.797} & \textbf{0.691} & \textbf{0.784} & \textbf{0.777} & \textbf{0.933} & \textbf{0.846} & \textbf{0.739} & \textbf{0.834} \\
\hline 
HRNet-W32 & $256\times192$ & - & 0.723 & 0.904 &  0.811 &  0.690 &  0.788 & 0.782 & 0.941 &  0.859 &  0.741 &  0.841\\
HRNet-W32 & $256\times192$ & * & 0.744 & 0.905 &  0.819 &  0.708 &  0.810 & 0.798 & 0.942 &  0.865 &  0.757 &  0.858\\
HRNet-W32 & $256\times192$ & \checkmark &  \textbf{0.750} & \textbf{0.906} & \textbf{0.820} & \textbf{0.715} & \textbf{0.817} & \textbf{0.802} & \textbf{0.942} & \textbf{0.865} & \textbf{0.761} & \textbf{0.861}\\
\hline
HRNet-W48 & $256\times192$ & - & 0.730 & 0.904 &  0.817 &  0.693 &  0.798 & 0.788 & 0.943 &  0.864 &  0.745 &  0.852 \\
HRNet-W48 & $256\times192$ & * & 0.751 & 0.906 &  0.822 &  0.715 &  0.818 & 0.804 & 0.943 &  0.867 &  0.762 &  0.864\\
HRNet-W48 & $256\times192$ & \checkmark &  \textbf{0.756} &  \textbf{0.906} &  \textbf{0.825} &  \textbf{0.718} &  \textbf{0.825} &  \textbf{0.806} &  \textbf{0.941} &  \textbf{0.868} &  \textbf{0.763} &  \textbf{0.869}\\
\hline
HRNet-W32 & $384\times288$ & - & 0.748 & 0.904 & 0.826 & 0.712 & 0.816 & 0.802 & 0.941 & 0.871 & 0.759 & 0.864  \\
HRNet-W32 & $384\times288$ & * & 0.758 & 0.906 &  0.825 &  0.720 &  0.827 & 0.809 & 0.943 &  0.869 &  0.767 &  0.871\\
HRNet-W32 & $384\times288$ & \checkmark &  \textbf{0.762} &  \textbf{0.905} &  \textbf{0.830} &  \textbf{0.725} &  \textbf{0.833} &  \textbf{0.812} &  \textbf{0.942} &  \textbf{0.873} &  \textbf{0.769} & \textbf{0.874}\\
\hline
HRNet-W48 & $384\times288$ & - & 0.753 & 0.907 & 0.823 & 0.712 & 0.823 & 0.804 & 0.941 & 0.867 & 0.759 & 0.869\\
HRNet-W48 & $384\times288$ & * & 0.763 & 0.908 &  0.829 &  0.723 &  0.834 & 0.812 & 0.942 &  0.871 &  0.767 &  0.876\\
HRNet-W48 & $384\times288$ &  \checkmark  &  \textbf{0.765} &  \textbf{0.907} &  \textbf{0.829} &  \textbf{0.724} &  \textbf{0.838} &  \textbf{0.814} &  \textbf{0.941} &  \textbf{0.871} &  \textbf{0.769} &  \textbf{0.878} \\
\hline
\end{tabular}
\end{center}
\end{table*}

\section{Human Pose Estimation}

In this section, we perform human pose estimation experiments to further demonstrate the effectiveness of the proposed quantization system for accurate semantic landmark localization. 

\subsection{Datasets}

We perform experiments on two popular human pose estimation datasets,
\begin{itemize}
\item \textbf{MPII}~\cite{andriluka20142d}: The MPII Human Pose dataset contains around 28,821 images with 40,522 person instances, in which 11,701 images for testing and the remaining 17,120 images for training. Following the experimental setup in ~\cite{sun2019deep}, we use 22,246 person instances for training and evaluate the performance on the MPII validation set with 2,958 person instances, which is a heldout subset of MPII training set. \\
\item \textbf{COCO}~\cite{lin2014microsoft}: The COCO dataset contains over 200,000 images and 250,000 person instances, in which each person instance is labeled with 17 keypoints. Following the experimental setup in~\cite{sun2019deep}, we evaluate the proposed method on the validation set with 5,000 images.
\end{itemize} 

\subsection{Implementation Details}

We utilize recent state-of-the-art heatmap regression method for human pose estimation, HRNet~\cite{sun2019deep}, as our baseline. Specifically, the proposed quantization system can be easily integrated into most heatmap regression models and we have made the source code of human pose estimation based on the HRNet baseline publicly available. For the MPII Human Pose dataset, we use the standard evaluation metric, head-normalized probability of correct keypoint or PCKh~\cite{andriluka20142d}. Specifically, a correct keypoint should fall within $\alpha*l$ pixels of the ground truth position, where $l$ indicates the normalization distance and $\alpha \in [0,1]$ is the matching threshold. For fair comparison, we apply two different matching thresholds, PCKh@0.5 and PCKh@0.1, where a smaller matching threshold, $\alpha=0.1$, indicates a more strict evaluation metric for accurate semantic landmark localization~\cite{sun2019deep}. For the COCO dataset, we use the standard evaluation metric, averaged precision (AP) and averaged recall (AR), where the object keypoint similarity or OKS is used as the similarity measure between the ground truth objects and the predicted objects~\cite{lin2014microsoft}.

\subsection{Results}

The experimental results on the MPII dataset are shown in Table~\ref{table:exp:pose:mpii}. Specifically, when using a coarse evaluation metric, PCKh@0.5, both the proposed method and the compensation method achieve comparable performance to the baseline method, suggesting that the quantization error is trivial in coarse semantic landmark localization; When using a more strict evaluation metric, PCKh@0.1, the compensation method, which can be seen as a special case of H3R with $k=2$, significantly improves the baseline, e.g., from $32.8$ to $37.7$, while the proposed method H3R further improves the performance from $37.7$ to $39.3$. The experimental results on the COCO dataset are shown in Table~\ref{table:exp:pose:coco}. Specifically, the proposed method clearly improves the averaged precision (AP) in different settings and the major improvements on AP come from: 1) a strict evaluation metric, e.g., AP$^{0.75}$; and 2) large/medium person instances, i.e., AP(M) and AP(L). Furthermore, we also find that the improvement decreases when increasing the input resolution, e.g., from $0.674$ to $0.720$ for $192\times128~(0.046{\uparrow})$, from $0.723$ to $0.750$ for $256\times192~(0.027{\uparrow})$, and from $0.748$ to $0.762$ for $384\times 288~(0.014{\uparrow})$. 

\section{Conclusion}

In this paper, we address the problem of sub-pixel localization for heatmap-based semantic landmark localization. We formally analyze quantization error in vanilla heatmap regression and propose a new quantization system via randomized rounding operation, which we prove is unbiased and lossless. Experiments on facial landmark localization and human pose estimation datasets demonstrate the effectiveness of the proposed quantization system for efficient and accurate sub-pixel localization. 

\section*{Acknowledgement}
Dr. Baosheng Yu is supported by ARC project FL-170100117.

\ifCLASSOPTIONcaptionsoff
  \newpage
\fi



\bibliographystyle{IEEEtran}
\bibliography{reference}

\begin{thebibliography}{10}
\providecommand{\url}[1]{#1}
\csname url@samestyle\endcsname
\providecommand{\newblock}{\relax}
\providecommand{\bibinfo}[2]{#2}
\providecommand{\BIBentrySTDinterwordspacing}{\spaceskip=0pt\relax}
\providecommand{\BIBentryALTinterwordstretchfactor}{4}
\providecommand{\BIBentryALTinterwordspacing}{\spaceskip=\fontdimen2\font plus
\BIBentryALTinterwordstretchfactor\fontdimen3\font minus
  \fontdimen4\font\relax}
\providecommand{\BIBforeignlanguage}[2]{{%
\expandafter\ifx\csname l@#1\endcsname\relax
\typeout{** WARNING: IEEEtran.bst: No hyphenation pattern has been}%
\typeout{** loaded for the language `#1'. Using the pattern for}%
\typeout{** the default language instead.}%
\else
\language=\csname l@#1\endcsname
\fi
#2}}
\providecommand{\BIBdecl}{\relax}
\BIBdecl

\bibitem{sun2013deep}
Y.~Sun, X.~Wang, and X.~Tang, ``Deep convolutional network cascade for facial
  point detection,'' in \emph{IEEE Conference on Computer Vision and Pattern
  Recognition (CVPR)}, 2013, pp. 3476--3483.

\bibitem{bulat2017far}
A.~Bulat and G.~Tzimiropoulos, ``How far are we from solving the 2d \& 3d face
  alignment problem?(and a dataset of 230,000 3d facial landmarks),'' in
  \emph{IEEE International Conference on Computer Vision (ICCV)}, 2017, pp.
  1021--1030.

\bibitem{sinha2016deephand}
A.~Sinha, C.~Choi, and K.~Ramani, ``Deephand: Robust hand pose estimation by
  completing a matrix imputed with deep features,'' in \emph{IEEE Conference on
  Computer Vision and Pattern Recognition (CVPR)}, June 2016.

\bibitem{iqbal2018hand}
U.~Iqbal, P.~Molchanov, T.~Breuel Juergen~Gall, and J.~Kautz, ``Hand pose
  estimation via latent 2.5 d heatmap regression,'' in \emph{European
  Conference on Computer Vision (ECCV)}, 2018, pp. 118--134.

\bibitem{toshev2014deeppose}
A.~Toshev and C.~Szegedy, ``Deeppose: Human pose estimation via deep neural
  networks,'' in \emph{IEEE Conference on Computer Vision and Pattern
  Recognition (CVPR)}, 2014, pp. 1653--1660.

\bibitem{newell2016stacked}
A.~Newell, K.~Yang, and J.~Deng, ``Stacked hourglass networks for human pose
  estimation,'' in \emph{European Conference on Computer Vision (ECCV)}, 2016,
  pp. 483--499.

\bibitem{ashutosh2008robotic}
\BIBentryALTinterwordspacing
A.~Saxena, J.~Driemeyer, and A.~Y. Ng, ``Robotic grasping of novel objects
  using vision,'' \emph{International Journal of Robotics Research (IJRR)},
  vol.~27, no.~2, pp. 157--173, 2008. [Online]. Available:
  \url{https://doi.org/10.1177/0278364907087172}
\BIBentrySTDinterwordspacing

\bibitem{sun2014deep}
Y.~Sun, Y.~Chen, X.~Wang, and X.~Tang, ``Deep learning face representation by
  joint identification-verification,'' in \emph{Neural Information Processing
  Systems (NIPS)}, 2014, pp. 1988--1996.

\bibitem{thies2016face2face}
J.~Thies, M.~Zollhofer, M.~Stamminger, C.~Theobalt, and M.~Nie{\ss}ner,
  ``Face2face: Real-time face capture and reenactment of rgb videos,'' in
  \emph{IEEE Conference on Computer Vision and Pattern Recognition (CVPR)},
  2016, pp. 2387--2395.

\bibitem{wang2019densefusion}
C.~Wang, D.~Xu, Y.~Zhu, R.~Mart{\'\i}n-Mart{\'\i}n, C.~Lu, L.~Fei-Fei, and
  S.~Savarese, ``Densefusion: 6d object pose estimation by iterative dense
  fusion,'' in \emph{IEEE Conference on Computer Vision and Pattern Recognition
  (CVPR)}, 2019.

\bibitem{deng2018arcface}
J.~Deng, J.~Guo, X.~Niannan, and S.~Zafeiriou, ``Arcface: Additive angular
  margin loss for deep face recognition,'' in \emph{IEEE Conference on Computer
  Vision and Pattern Recognition (CVPR)}, 2019.

\bibitem{zhao2017spindle}
H.~Zhao, M.~Tian, S.~Sun, J.~Shao, J.~Yan, S.~Yi, X.~Wang, and X.~Tang,
  ``Spindle net: Person re-identification with human body region guided feature
  decomposition and fusion,'' in \emph{IEEE Conference on Computer Vision and
  Pattern Recognition (CVPR)}, 2017, pp. 1077--1085.

\bibitem{zheng2019pose}
L.~Zheng, Y.~Huang, H.~Lu, and Y.~Yang, ``Pose-invariant embedding for deep
  person re-identification,'' \emph{IEEE Transactions on Image Processing
  (TIP)}, vol.~28, no.~9, pp. 4500--4509, 2019.

\bibitem{chen2017makeup}
Y.-C. Chen, X.~Shen, and J.~Jia, ``Makeup-go: Blind reversion of portrait
  edit,'' in \emph{IEEE International Conference on Computer Vision (ICCV)},
  Oct 2017.

\bibitem{chang2018pairedcyclegan}
H.~Chang, J.~Lu, F.~Yu, and A.~Finkelstein, ``Pairedcyclegan: Asymmetric style
  transfer for applying and removing makeup,'' in \emph{IEEE Conference on
  Computer Vision and Pattern Recognition (CVPR)}, 2018, pp. 40--48.

\bibitem{cao20133d}
C.~Cao, Y.~Weng, S.~Lin, and K.~Zhou, ``3d shape regression for real-time
  facial animation,'' \emph{ACM Transactions on Graphics (TOG)}, vol.~32,
  no.~4, pp. 1--10, 2013.

\bibitem{cao2014displaced}
C.~Cao, Q.~Hou, and K.~Zhou, ``Displaced dynamic expression regression for
  real-time facial tracking and animation,'' \emph{ACM Transactions on Graphics
  (TOG)}, vol.~33, no.~4, pp. 1--10, 2014.

\bibitem{thies2015real}
J.~Thies, M.~Zollh{\"o}fer, M.~Nie{\ss}ner, L.~Valgaerts, M.~Stamminger, and
  C.~Theobalt, ``Real-time expression transfer for facial reenactment.''
  \emph{ACM Transactions on Graphics (TOG)}, vol.~34, no.~6, pp. 183--1, 2015.

\bibitem{tompson2014joint}
J.~J. Tompson, A.~Jain, Y.~LeCun, and C.~Bregler, ``Joint training of a
  convolutional network and a graphical model for human pose estimation,'' in
  \emph{Neural Information Processing Systems (NIPS)}, 2014, pp. 1799--1807.

\bibitem{nibali2018numerical}
A.~Nibali, Z.~He, S.~Morgan, and L.~Prendergast, ``Numerical coordinate
  regression with convolutional neural networks,'' \emph{arXiv preprint
  arXiv:1801.07372}, 2018.

\bibitem{wang2019adaptive}
X.~Wang, L.~Bo, and L.~Fuxin, ``Adaptive wing loss for robust face alignment
  via heatmap regression,'' in \emph{IEEE International Conference on Computer
  Vision (ICCV)}, 2019, pp. 6971--6981.

\bibitem{sun2019deep}
K.~Sun, B.~Xiao, D.~Liu, and J.~Wang, ``Deep high-resolution representation
  learning for human pose estimation,'' in \emph{IEEE Conference on Computer
  Vision and Pattern Recognition (CVPR)}, 2019, pp. 5693--5703.

\bibitem{sun2019high}
J.~Wang, K.~Sun, T.~Cheng, B.~Jiang, C.~Deng, Y.~Zhao, D.~Liu, Y.~Mu, M.~Tan,
  X.~Wang, W.~Liu, and B.~Xiao, ``Deep high-resolution representation learning
  for visual recognition,'' \emph{IEEE Transactions on Pattern Analysis and
  Machine Intelligence (TPAMI)}, 2020.

\bibitem{tai2019towards}
Y.~Tai, Y.~Liang, X.~Liu, L.~Duan, J.~Li, C.~Wang, F.~Huang, and Y.~Chen,
  ``Towards highly accurate and stable face alignment for high-resolution
  videos,'' in \emph{AAAI Conference on Artificial Intelligence (AAAI)},
  vol.~33, 2019, pp. 8893--8900.

\bibitem{li2019rethinking}
W.~Li, Z.~Wang, B.~Yin, Q.~Peng, Y.~Du, T.~Xiao, G.~Yu, H.~Lu, Y.~Wei, and
  J.~Sun, ``Rethinking on multi-stage networks for human pose estimation,''
  \emph{arXiv preprint arXiv:1901.00148}, 2019.

\bibitem{zhang2019distribution}
F.~Zhang, X.~Zhu, H.~Dai, M.~Ye, and C.~Zhu, ``Distribution-aware coordinate
  representation for human pose estimation,'' in \emph{IEEE Conference on
  Computer Vision and Pattern Recognition (CVPR)}, 2020, pp. 7093--7102.

\bibitem{raghavan1987randomized}
P.~Raghavan and C.~D. Tompson, ``Randomized rounding: a technique for provably
  good algorithms and algorithmic proofs,'' \emph{Combinatorica}, vol.~7,
  no.~4, pp. 365--374, 1987.

\bibitem{korte2012combinatorial}
B.~Korte, J.~Vygen, B.~Korte, and J.~Vygen, \emph{Combinatorial
  optimization}.\hskip 1em plus 0.5em minus 0.4em\relax Springer, 2012, vol.~2.

\bibitem{cao2014face}
X.~Cao, Y.~Wei, F.~Wen, and J.~Sun, ``Face alignment by explicit shape
  regression,'' \emph{International Journal of Computer Vision (IJCV)}, vol.
  107, no.~2, pp. 177--190, 2014.

\bibitem{zhu2012face}
X.~Zhu and D.~Ramanan, ``Face detection, pose estimation, and landmark
  localization in the wild,'' in \emph{IEEE Conference on Computer Vision and
  Pattern Recognition (CVPR)}, 2012, pp. 2879--2886.

\bibitem{xiong2013supervised}
X.~Xiong and F.~De~la Torre, ``Supervised descent method and its applications
  to face alignment,'' in \emph{IEEE Conference on Computer Vision and Pattern
  Recognition (CVPR)}, 2013, pp. 532--539.

\bibitem{ren2014face}
S.~Ren, X.~Cao, Y.~Wei, and J.~Sun, ``Face alignment at 3000 fps via regressing
  local binary features,'' in \emph{IEEE Conference on Computer Vision and
  Pattern Recognition (CVPR)}, 2014, pp. 1685--1692.

\bibitem{kazemi2014one}
V.~Kazemi and J.~Sullivan, ``One millisecond face alignment with an ensemble of
  regression trees,'' in \emph{IEEE Conference on Computer Vision and Pattern
  Recognition (CVPR)}, 2014, pp. 1867--1874.

\bibitem{burgos2013robust}
X.~P. Burgos-Artizzu, P.~Perona, and P.~Doll{\'a}r, ``Robust face landmark
  estimation under occlusion,'' in \emph{IEEE International Conference on
  Computer Vision (ICCV)}, 2013, pp. 1513--1520.

\bibitem{zhang2014coarse}
J.~Zhang, S.~Shan, M.~Kan, and X.~Chen, ``Coarse-to-fine auto-encoder networks
  (cfan) for real-time face alignment,'' in \emph{European Conference on
  Computer Vision (ECCV)}, 2014, pp. 1--16.

\bibitem{zhu2015face}
S.~Zhu, C.~Li, C.~Change~Loy, and X.~Tang, ``Face alignment by coarse-to-fine
  shape searching,'' in \emph{IEEE Conference on Computer Vision and Pattern
  Recognition (CVPR)}, 2015, pp. 4998--5006.

\bibitem{carreira2016human}
J.~Carreira, P.~Agrawal, K.~Fragkiadaki, and J.~Malik, ``Human pose estimation
  with iterative error feedback,'' in \emph{IEEE Conference on Computer Vision
  and Pattern Recognition (CVPR)}, 2016, pp. 4733--4742.

\bibitem{belagiannis2017recurrent}
V.~Belagiannis and A.~Zisserman, ``Recurrent human pose estimation,'' in
  \emph{IEEE International Conference on Automatic Face \& Gesture Recognition
  (FG)}, 2017, pp. 468--475.

\bibitem{belhumeur2013localizing}
P.~N. Belhumeur, D.~W. Jacobs, D.~J. Kriegman, and N.~Kumar, ``Localizing parts
  of faces using a consensus of exemplars,'' \emph{IEEE Transactions on Pattern
  Analysis and Machine Intelligence (TPAMI)}, vol.~35, no.~12, pp. 2930--2940,
  2013.

\bibitem{miao2018direct}
X.~Miao, X.~Zhen, X.~Liu, C.~Deng, V.~Athitsos, and H.~Huang, ``Direct shape
  regression networks for end-to-end face alignment,'' in \emph{IEEE Conference
  on Computer Vision and Pattern Recognition (CVPR)}, 2018, pp. 5040--5049.

\bibitem{zhang2014facial}
Z.~Zhang, P.~Luo, C.~C. Loy, and X.~Tang, ``Facial landmark detection by deep
  multi-task learning,'' in \emph{European Conference on Computer Vision
  (ECCV)}, 2014, pp. 94--108.

\bibitem{zhang2016joint}
K.~Zhang, Z.~Zhang, Z.~Li, and Y.~Qiao, ``Joint face detection and alignment
  using multitask cascaded convolutional networks,'' \emph{IEEE Signal
  Processing Letters (SPL)}, vol.~23, no.~10, pp. 1499--1503, 2016.

\bibitem{ranjan2017hyperface}
R.~Ranjan, V.~M. Patel, and R.~Chellappa, ``Hyperface: A deep multi-task
  learning framework for face detection, landmark localization, pose
  estimation, and gender recognition,'' \emph{IEEE Transactions on Pattern
  Analysis and Machine Intelligence (TPAMI)}, vol.~41, no.~1, pp. 121--135,
  2017.

\bibitem{girshick2015fast}
R.~Girshick, ``Fast r-cnn,'' in \emph{IEEE International Conference on Computer
  Vision (ICCV)}, 2015, pp. 1440--1448.

\bibitem{feng2018wing}
Z.-H. Feng, J.~Kittler, M.~Awais, P.~Huber, and X.-J. Wu, ``Wing loss for
  robust facial landmark localisation with convolutional neural networks,'' in
  \emph{IEEE Conference on Computer Vision and Pattern Recognition (CVPR)},
  2018, pp. 2235--2245.

\bibitem{bulat2017binarized}
A.~Bulat and G.~Tzimiropoulos, ``Binarized convolutional landmark localizers
  for human pose estimation and face alignment with limited resources,'' in
  \emph{IEEE International Conference on Computer Vision (ICCV)}, 2017, pp.
  3706--3714.

\bibitem{merget2018robust}
D.~Merget, M.~Rock, and G.~Rigoll, ``Robust facial landmark detection via a
  fully-convolutional local-global context network,'' in \emph{IEEE Conference
  on Computer Vision and Pattern Recognition (CVPR)}, 2018, pp. 781--790.

\bibitem{pfister2015flowing}
T.~Pfister, J.~Charles, and A.~Zisserman, ``Flowing convnets for human pose
  estimation in videos,'' in \emph{IEEE International Conference on Computer
  Vision (ICCV)}, 2015, pp. 1913--1921.

\bibitem{pishchulin2016deepcut}
L.~Pishchulin, E.~Insafutdinov, S.~Tang, B.~Andres, M.~Andriluka, P.~V. Gehler,
  and B.~Schiele, ``Deepcut: Joint subset partition and labeling for multi
  person pose estimation,'' in \emph{IEEE Conference on Computer Vision and
  Pattern Recognition (CVPR)}, 2016, pp. 4929--4937.

\bibitem{newell2017associative}
A.~Newell, Z.~Huang, and J.~Deng, ``Associative embedding: End-to-end learning
  for joint detection and grouping,'' in \emph{Neural Information Processing
  Systems (NIPS)}, 2017, pp. 2277--2287.

\bibitem{yang2017learning}
W.~Yang, S.~Li, W.~Ouyang, H.~Li, and X.~Wang, ``Learning feature pyramids for
  human pose estimation,'' in \emph{IEEE International Conference on Computer
  Vision (ICCV)}, 2017, pp. 1281--1290.

\bibitem{papandreou2018personlab}
G.~Papandreou, T.~Zhu, L.-C. Chen, S.~Gidaris, J.~Tompson, and K.~Murphy,
  ``Personlab: Person pose estimation and instance segmentation with a
  bottom-up, part-based, geometric embedding model,'' in \emph{European
  Conference on Computer Vision (ECCV)}, 2018, pp. 269--286.

\bibitem{wei2016convolutional}
S.-E. Wei, V.~Ramakrishna, T.~Kanade, and Y.~Sheikh, ``Convolutional pose
  machines,'' in \emph{IEEE Conference on Computer Vision and Pattern
  Recognition (CVPR)}, 2016, pp. 4724--4732.

\bibitem{cao2017realtime}
Z.~Cao, T.~Simon, S.-E. Wei, and Y.~Sheikh, ``Realtime multi-person 2d pose
  estimation using part affinity fields,'' in \emph{IEEE Conference on Computer
  Vision and Pattern Recognition (CVPR)}, 2017, pp. 7291--7299.

\bibitem{chen2018cascaded}
Y.~Chen, Z.~Wang, Y.~Peng, Z.~Zhang, G.~Yu, and J.~Sun, ``Cascaded pyramid
  network for multi-person pose estimation,'' in \emph{IEEE Conference on
  Computer Vision and Pattern Recognition (CVPR)}, 2018, pp. 7103--7112.

\bibitem{papandreou2017towards}
G.~Papandreou, T.~Zhu, N.~Kanazawa, A.~Toshev, J.~Tompson, C.~Bregler, and
  K.~Murphy, ``Towards accurate multi-person pose estimation in the wild,'' in
  \emph{IEEE Conference on Computer Vision and Pattern Recognition (CVPR)},
  2017, pp. 4903--4911.

\bibitem{sun2018integral}
X.~Sun, B.~Xiao, F.~Wei, S.~Liang, and Y.~Wei, ``Integral human pose
  regression,'' in \emph{European Conference on Computer Vision (ECCV)}, 2018,
  pp. 529--545.

\bibitem{luvizon20182d}
D.~C. Luvizon, D.~Picard, and H.~Tabia, ``2d/3d pose estimation and action
  recognition using multitask deep learning,'' in \emph{IEEE Conference on
  Computer Vision and Pattern Recognition (CVPR)}, 2018, pp. 5137--5146.

\bibitem{luvizon2019human}
D.~C. Luvizon, H.~Tabia, and D.~Picard, ``Human pose regression by combining
  indirect part detection and contextual information,'' \emph{Computers \&
  Graphics}, vol.~85, pp. 15--22, 2019.

\bibitem{yi2016lift}
K.~M. Yi, E.~Trulls, V.~Lepetit, and P.~Fua, ``Lift: Learned invariant feature
  transform,'' in \emph{European Conference on Computer Vision (ECCV)}, 2016,
  pp. 467--483.

\bibitem{levine2016end}
S.~Levine, C.~Finn, T.~Darrell, and P.~Abbeel, ``End-to-end training of deep
  visuomotor policies,'' \emph{Journal of Machine Learning Research (JMLR)},
  vol.~17, no.~1, pp. 1334--1373, 2016.

\bibitem{thewlis2017unsupervised}
J.~Thewlis, H.~Bilen, and A.~Vedaldi, ``Unsupervised learning of object
  landmarks by factorized spatial embeddings,'' in \emph{IEEE International
  Conference on Computer Vision (ICCV)}, 2017, pp. 5916--5925.

\bibitem{wu2017leveraging}
W.~Wu and S.~Yang, ``Leveraging intra and inter-dataset variations for robust
  face alignment,'' in \emph{IEEE Conference on Computer Vision and Pattern
  Recognition Workshops (CVPRW)}, 2017, pp. 150--159.

\bibitem{wu2018look}
W.~Wu, C.~Qian, S.~Yang, Q.~Wang, Y.~Cai, and Q.~Zhou, ``Look at boundary: A
  boundary-aware face alignment algorithm,'' in \emph{IEEE Conference on
  Computer Vision and Pattern Recognition (CVPR)}, 2018, pp. 2129--2138.

\bibitem{valle2019face}
R.~Valle, J.~M. Buenaposada, A.~Vald{\'e}s, and L.~Baumela, ``Face alignment
  using a 3d deeply-initialized ensemble of regression trees,'' \emph{Computer
  Vision and Image Understanding (CVIU)}, vol. 189, p. 102846, 2019.

\bibitem{Dapogny_2019_ICCV}
A.~Dapogny, K.~Bailly, and M.~Cord, ``Decafa: Deep convolutional cascade for
  face alignment in the wild,'' in \emph{IEEE International Conference on
  Computer Vision (ICCV)}, October 2019.

\bibitem{qian2019aggregation}
S.~Qian, K.~Sun, W.~Wu, C.~Qian, and J.~Jia, ``Aggregation via separation:
  Boosting facial landmark detector with semi-supervised style translation,''
  in \emph{IEEE International Conference on Computer Vision (ICCV)}, 2019, pp.
  10\,153--10\,163.

\bibitem{kumar2020luvli}
A.~Kumar, T.~K. Marks, W.~Mou, Y.~Wang, M.~Jones, A.~Cherian, T.~Koike-Akino,
  X.~Liu, and C.~Feng, ``Luvli face alignment: Estimating landmarks' location,
  uncertainty, and visibility likelihood,'' in \emph{IEEE Conference on
  Computer Vision and Pattern Recognition (CVPR)}, 2020, pp. 8236--8246.

\bibitem{dong2018style}
X.~Dong, Y.~Yan, W.~Ouyang, and Y.~Yang, ``Style aggregated network for facial
  landmark detection,'' in \emph{IEEE Conference on Computer Vision and Pattern
  Recognition (CVPR)}, 2018, pp. 379--388.

\bibitem{kowalski2017deep}
M.~Kowalski, J.~Naruniec, and T.~Trzcinski, ``Deep alignment network: A
  convolutional neural network for robust face alignment,'' in \emph{IEEE
  Conference on Computer Vision and Pattern Recognition Workshops (CVPRW)},
  2017, pp. 88--97.

\bibitem{yang2017stacked}
J.~Yang, Q.~Liu, and K.~Zhang, ``Stacked hourglass network for robust facial
  landmark localisation,'' in \emph{IEEE Conference on Computer Vision and
  Pattern Recognition Workshops (CVPRW)}, 2017, pp. 79--87.

\bibitem{valle2018deeply}
R.~Valle, J.~M. Buenaposada, A.~Vald{\'e}s, and L.~Baumela, ``A
  deeply-initialized coarse-to-fine ensemble of regression trees for face
  alignment,'' in \emph{European Conference on Computer Vision (ECCV)}, 2018,
  pp. 585--601.

\bibitem{Zou_2019_ICCV}
X.~Zou, S.~Zhong, L.~Yan, X.~Zhao, J.~Zhou, and Y.~Wu, ``Learning robust facial
  landmark detection via hierarchical structured ensemble,'' in \emph{IEEE
  International Conference on Computer Vision (ICCV)}, October 2019.

\bibitem{yang2016wider}
S.~Yang, P.~Luo, C.-C. Loy, and X.~Tang, ``Wider face: A face detection
  benchmark,'' in \emph{IEEE Conference on Computer Vision and Pattern
  Recognition (CVPR)}, 2016, pp. 5525--5533.

\bibitem{sagonas2013300}
C.~Sagonas, G.~Tzimiropoulos, S.~Zafeiriou, and M.~Pantic, ``300 faces
  in-the-wild challenge: The first facial landmark localization challenge,'' in
  \emph{IEEE International Conference on Computer Vision Workshops (ICCVW)},
  2013, pp. 397--403.

\bibitem{le2012interactive}
V.~Le, J.~Brandt, Z.~Lin, L.~Bourdev, and T.~S. Huang, ``Interactive facial
  feature localization,'' in \emph{European Conference on Computer Vision
  (ECCV)}, 2012, pp. 679--692.

\bibitem{koestinger2011annotated}
M.~Koestinger, P.~Wohlhart, P.~M. Roth, and H.~Bischof, ``Annotated facial
  landmarks in the wild: A large-scale, real-world database for facial landmark
  localization,'' in \emph{IEEE International Conference on Computer Vision
  Workshops (ICCVW)}, 2011, pp. 2144--2151.

\bibitem{NIPS2019_9015}
A.~Paszke, S.~Gross, F.~Massa, A.~Lerer, J.~Bradbury, G.~Chanan, T.~Killeen,
  Z.~Lin, N.~Gimelshein, L.~Antiga, A.~Desmaison, A.~Kopf, E.~Yang, Z.~DeVito,
  M.~Raison, A.~Tejani, S.~Chilamkurthy, B.~Steiner, L.~Fang, J.~Bai, and
  S.~Chintala, ``Pytorch: An imperative style, high-performance deep learning
  library,'' in \emph{Neural Information Processing Systems (NeurIPS)},
  H.~Wallach, H.~Larochelle, A.~Beygelzimer, F.~d~Alch\'{e}-Buc, E.~Fox, and
  R.~Garnett, Eds.\hskip 1em plus 0.5em minus 0.4em\relax Curran Associates,
  Inc., 2019, pp. 8024--8035.

\bibitem{he2016deep}
K.~He, X.~Zhang, S.~Ren, and J.~Sun, ``Deep residual learning for image
  recognition,'' in \emph{IEEE Conference on Computer Vision and Pattern
  Recognition (CVPR)}, 2016, pp. 770--778.

\bibitem{ronneberger2015u}
O.~Ronneberger, P.~Fischer, and T.~Brox, ``U-net: Convolutional networks for
  biomedical image segmentation,'' in \emph{International Conference on Medical
  Image Computing and Computer-Assisted Intervention (MICCAI)}.\hskip 1em plus
  0.5em minus 0.4em\relax Springer, 2015, pp. 234--241.

\bibitem{zhong2017random}
Z.~Zhong, L.~Zheng, G.~Kang, S.~Li, and Y.~Yang, ``Random erasing data
  augmentation,'' \emph{AAAI Conference on Artificial Intelligence (AAAI)},
  2020.

\bibitem{deng2009imagenet}
J.~Deng, W.~Dong, R.~Socher, L.-J. Li, K.~Li, and L.~Fei-Fei, ``Imagenet: A
  large-scale hierarchical image database,'' in \emph{IEEE Conference on
  Computer Vision and Pattern Recognition (CVPR)}, 2009, pp. 248--255.

\bibitem{kingma2015adam}
D.~P. Kingma and J.~Ba, ``Adam: A method for stochastic optimization,'' in
  \emph{International Conference on Learning Representations (ICLR)}, 2015.

\bibitem{jacek2020high}
J.~Naruniec, L.~Helminger, C.~Schroers, and R.~Weber, ``High-resolution neural
  face swapping for visual effects,'' \emph{Eurographics Symposium on
  Rendering}, vol.~39, no.~4, 2020.

\bibitem{andriluka20142d}
M.~Andriluka, L.~Pishchulin, P.~Gehler, and B.~Schiele, ``2d human pose
  estimation: New benchmark and state of the art analysis,'' in \emph{IEEE
  Conference on Computer Vision and Pattern Recognition (CVPR)}, 2014, pp.
  3686--3693.

\bibitem{lin2014microsoft}
T.-Y. Lin, M.~Maire, S.~Belongie, J.~Hays, P.~Perona, D.~Ramanan,
  P.~Doll{\'a}r, and C.~L. Zitnick, ``Microsoft coco: Common objects in
  context,'' in \emph{European Conference on Computer Vision (ECCV)}, 2014, pp.
  740--755.

\end{thebibliography}
%

%

\begin{IEEEbiography}[{\includegraphics[width=1in,height=1.25in,clip,keepaspectratio]{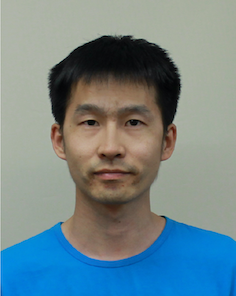}}]{Baosheng Yu} received a B.E. from the University of Science and Technology of China in 2014, and a Ph.D. from The University of Sydney in 2019. He is currently a Research Fellow in the School of Computer Science and the Faculty of Engineering at The University of Sydney, NSW, Australia. His research interests include machine learning, computer vision, and deep learning.
\end{IEEEbiography}

\begin{IEEEbiography}[{\includegraphics[width=1in,height=1.25in,clip]{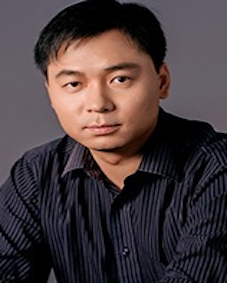}}]{Dacheng Tao (F’15)} is the President of the JD Explore Academy and a Senior Vice President of JD.com. He is also an advisor and chief scientist of the digital science institute in the University of Sydney. He mainly applies statistics and mathematics to artificial intelligence and data science, and his research is detailed in one monograph and over 200 publications in prestigious journals and proceedings at leading conferences. He received the 2015 Australian Scopus-Eureka Prize, the 2018 IEEE ICDM Research Contributions Award, and the 2021 IEEE Computer Society McCluskey Technical Achievement Award. He is a fellow of the Australian Academy of Science, AAAS, and ACM.
\end{IEEEbiography}




\newpage
\appendices
\section{Proofs of Theorem 1 and Theorem 2}
\setcounter{theorem}{0}

\begin{theorem}
Given an unbiased quantization system defined by the encode operation in~\eqref{eq:quantization} and the decode operation in~\eqref{eq:argmax:unbiased}, we then have that the quantization error tightly upper bounded, i.e.,
\begin{equation}
    \| \boldsymbol{x}_i^p - \boldsymbol{x}_i^g \|_2 \leq \sqrt{2}s/2, \nonumber
\end{equation}
where $s>1$ indicates the downsampling stride of the heatmap.

\begin{proof}
Given the ground truth numerical coordinate $\boldsymbol{x}_i^g = \left(x_i^g, y_i^g\right)$, the predicted numerical coordinate $\boldsymbol{x}_i^p = \left(x_i^p, y_i^p\right)$, and the downsampling stride of the heatmap $s>1$, if there is no heatmap error, we then have 
\begin{equation}
\boldsymbol{h}_i^p(\boldsymbol{x}) = \boldsymbol{h}_i^g(\boldsymbol{x}), \nonumber
\end{equation}
where $\boldsymbol{h}_i^p(\boldsymbol{x})$ and  $\boldsymbol{h}_i^g(\boldsymbol{x})$ indicate the ground truth heatmap and the predicted heatmap, respectively. Therefore, according to the decode operation in~\eqref{eq:argmax:unbiased}, we have the predicted numerical coordinate as 
\begin{equation}
x_i^p/s =
\begin{cases}
\lfloor x_i^g/s \rfloor + t-0.5 & \quad \text{if }~\epsilon_x < t, \\ \nonumber
\lfloor x_i^g/s \rfloor + t+0.5 & \quad \text{otherwise}.
\end{cases}
\end{equation}
\begin{equation}
y_i^p/s =
\begin{cases}
\lfloor y_i^g/s \rfloor + t-0.5 & \quad \text{if }~\epsilon_y < t, \\ \nonumber
\lfloor y_i^g/s \rfloor + t+0.5 & \quad \text{otherwise}. 
\end{cases}
\end{equation}
where $\epsilon_x = x_i^g/s - \lfloor x_i^g/s \rfloor$ and $\epsilon_y = y_i^g/s - \lfloor y_i^g/s \rfloor $. The quantization error of vanilla quantization system then can be evaluated as follows:
\begin{equation}
|x_i^p/s - x_i^g/s| = 
\begin{cases}
| t-\epsilon_x-0.5 | & \quad \text{if }~\epsilon_x < t, \\ \nonumber
| t-\epsilon_x+0.5 | & \quad \text{otherwise}.
\end{cases}
\end{equation}
\begin{equation}
|y_i^p/s - y_i^g/s| = 
\begin{cases}
| t-\epsilon_y-0.5 | & \quad \text{if }~\epsilon_y < t, \\ \nonumber
| t-\epsilon_y+0.5 | & \quad \text{otherwise}.
\end{cases}
\end{equation}
The maximum quantization error $|x_i^p-x_i^g| = s/2$ is achieved when $\epsilon_x = t$. Similarly, we have the maximum quantization error $|y_i^p-y_i^g| = s/2$ is achieved with $\epsilon_y = t$. Considering that $x_i^p$ and $y_i^p$ are linearly independent variables, we thus have
\begin{equation}
\begin{split}
    \|\boldsymbol{x}_i^p - \boldsymbol{x}_i^g\|_2 &= \sqrt{\left(x_i^p-x_i^g\right)^2 + \left(y_i^p-y_i^g\right)^2}\nonumber \leq \sqrt{2}s/2. \nonumber
\end{split}
\end{equation}
The maximum quantization error is achieved with $\epsilon_x = \epsilon_y= t$.  That is, the quantization error in vanilla quantization system is tightly upper bounded by $\sqrt{2}s/2$. \\
\end{proof}

\end{theorem}

\begin{theorem}
Given the encode operation in~\eqref{eq:quantization:random} and the decode operation in~\eqref{eq:decode:random}, we then have that the 1) encode operation is unbiased; and 2) quantization system is lossless, i.e., there is no quantization error.

\begin{proof}
Given the ground truth numerical coordinate $\boldsymbol{x}_i^g = \left(x_i^g, y_i^g\right)$, the predicted numerical coordinate $\boldsymbol{x}_i^p = \left(x_i^p, y_i^p\right)$, and the downsampling stride of the heatmap $s>1$, we then have
\begin{equation}
\begin{split}
    \mathbb{E}\left( \boldsymbol{q}(x_i^g,s) \right) &=  \mathbb{E}\left( P\left\{\epsilon_x < t \right\}\lfloor x_i^g/s \rfloor + P\left\{\epsilon_x \geq t\right\} (\lfloor x_i^g/s \rfloor+1) \right)\\ \nonumber
    &= \lfloor x_i^g/s \rfloor (1-\epsilon_x) + (\lfloor x_i^g/s \rfloor + 1)\epsilon_x \\
    &= x_i^g/s
\end{split}
\end{equation}
Similarly, we have $\mathbb{E}\left( \boldsymbol{q}(y_i^g,s) \right)=y_i^g/s$. Considering that $x_i^p$ and $y_i^p$ are linearly independent variables, we thus have
\begin{equation}
\begin{split}
 \mathbb{E}\left( \boldsymbol{q}(\boldsymbol{x}_i^g,s) \right) &=  \left(\mathbb{E}\left( \boldsymbol{q}(y_i^g,s) \right),  \mathbb{E}\left( \boldsymbol{q}(x_i^g,s) \right)\right) \\ \nonumber
 &= (x_i^g/s, y_i^g/s).
\end{split}
\end{equation}
Therefore, the encode operation in \eqref{eq:quantization:random}, i.e., \textbf{random-round}, is an unbiased encode operation for heatmap regression. 

We then prove that the quantization system is losses as follows. For the decode operation in~\eqref{eq:decode:random},  if there is no heatmap error, we then have
\begin{equation}
\begin{split}
P\{\boldsymbol{x}_i^p/s=(\lfloor x_i^g/s \rfloor, \lfloor y_i^g/s \rfloor)\} &= (1-\epsilon_x)(1-\epsilon_y), \\ \nonumber
P\{\boldsymbol{x}_i^p/s=(\lfloor x_i^g/s \rfloor+1, \lfloor y_i^g/s \rfloor)\} &= \epsilon_x(1-\epsilon_y), \\
P\{\boldsymbol{x}_i^p/s=(\lfloor x_i^g/s \rfloor, \lfloor y_i^g/s \rfloor+1)\} &= (1-\epsilon_x)\epsilon_y, \\
P\{\boldsymbol{x}_i^p/s=(\lfloor x_i^g/s \rfloor+1, \lfloor y_i^g/s \rfloor+1)\} &= \epsilon_x\epsilon_y. \\
\end{split}
\end{equation}
We can reconstruct the fractional part of $\boldsymbol{x}_i^g$, i.e.,
\begin{equation}
\begin{split}
  \left(x_i^p/s, y_i^p/s\right) =& \sum\limits_{\boldsymbol{x}_i^p} P\{\boldsymbol{x}_i^p\}\boldsymbol{x}_i^p/s \\ \nonumber
  =~&(\lfloor x_i^g/s \rfloor, \lfloor y_i^g/s \rfloor) * (1-\epsilon_x)(1-\epsilon_y) \\
  &+(\lfloor x_i^g/s \rfloor+1, \lfloor y_i^g/s \rfloor) * \epsilon_x(1-\epsilon_y)\\
  &+(\lfloor x_i^g/s \rfloor, \lfloor y_i^g/s \rfloor+1) *  (1-\epsilon_x)\epsilon_y\\
  &+ (\lfloor x_i^g/s \rfloor+1, \lfloor y_i^g/s \rfloor+1) * \epsilon_x\epsilon_y\\
  =& \left(x_i^g/s, y_i^g/s\right).
\end{split}
\end{equation}
That is, $\left(x_i^p, y_i^p\right)= \left(x_i^g, y_i^g\right)$, i.e., there is no quantization error.\\
\end{proof}

\end{theorem}

\section{Experiments}

In this section, we provide additional experimental results on facial landmark detection and human pose estimation.

\subsection{Facial Landmark Detection}

\begin{table}[ht]
\caption{The influence of different numbers of training samples when using different input resolutions. In each cell, the first number indicates the performance of baseline method and the second number indicates the performance of the proposed method.}
\label{table:exp:ablation:sample-size}
\begin{center}
\begin{tabular}{|c|c|c|c|}
\hline
\multirow{2}{*}{\#Samples} & \multicolumn{3}{c|}{NME (\%)} \\
\cline{2-4} & $256\times256$ & $128\times128$ & $64\times64$ \\
\hline
256 & 5.67/5.46 & 6.02/5.47 &  7.74/6.20 \\
\hline
1024 & 4.79/4.63 & 5.27/4.77 & 7.13/5.43 \\
\hline
4096 & 4.14/3.97 & 4.81/4.17 &  6.57/4.76 \\
\hline
7500 & 4.00/3.81 & 4.62/3.99 &  6.50/4.62 \\
\hline 
\end{tabular}
\end{center}
\end{table}

\begin{table*}[!ht]
\caption{Comparison of different backbone networks and feature maps on WFLW dataset.}
\label{table:appendix:backbone:wflw}
\begin{center}
\begin{tabular}{|c|c|c|c|c|c|c|c|c|c|c|c|}
\hline 
\multirow{2}{*}{Backbone} & \multirow{2}{*}{Input} & \multirow{2}{*}{Heatmap} & \multirow{2}{*}{FLOPs} & \multirow{2}{*}{\#Params} & \multicolumn{7}{c|}{NME (\%), Inter-ocular} \\
\cline{6-12} & & & & & \textbf{test} & \textbf{pose} &  \textbf{expression} &  \textbf{illumination} &  \textbf{make-up} & \textbf{occlusion} & \textbf{blur} \\
\hline
HRNet-W18 & $256\times256$ & $64\times64$ & 4.84G & 9.69M & 3.81 & 6.45 & 4.07 & 3.70 & 3.66 & 4.48 & 4.30 \\
HRNet-W18  & $256\times256$ & $256\times256$ & 4.98G & 9.69M & 3.91 & 6.83 & 4.09 & 3.82 & 3.70 & 4.59 & 4.41  \\
U-Net & $256\times256$ & $256\times256$ & 60.55G & 31.46M & 4.93 & 9.30 & 5.08 & 4.74 &  4.82 & 6.43 & 5.72 \\
\hline
HRNet-W18 & $128\times128$ &  $32\times32$ & 1.21G & 9.69M & 3.99 & 6.78 & 4.26 & 3.89 & 3.84 & 4.65 & 4.46 \\
HRNet-W18 & $128\times128$ &  $128\times128$ & 1.24G  & 9.69M & 4.06 & 6.89 & 4.41 & 3.97 & 3.95 & 4.78 & 4.52 \\
U-Net & $128\times128$ & $128\times128$ & 15.14G & 31.46M & 4.17 & 7.10 & 4.45 & 4.06 & 4.03 & 5.00 & 4.73  \\ 
\hline
HRNet-W18 & $64\times64$ &  $16\times16$ & 0.30G & 9.69M& 4.64 & 7.77 & 5.05 & 4.52 & 4.59 & 5.31 & 4.96 \\
HRNet-W18 & $64\times64$ & $64\times64$ &  0.31G & 9.69M& 4.61 & 7.70 & 5.00 & 4.44 & 4.58 & 5.28 & 4.94 \\
U-Net & $64\times64$ & $64\times64$ & 3.79G & 31.46M& 4.37 & 7.18 & 4.69 & 4.26 & 4.30 & 5.12 & 4.80 \\
\hline
\end{tabular}
\end{center}
\end{table*}

\textbf{The influence of different numbers of training samples}. The proposed quantization system does not rely on any assumption about the number of training samples, and is lossless for heatmap regression if there is no heatmap error. However, heatmap prediction performance will be influenced by the number of training samples: increasing the number of training samples improves the model generalizability from the learning theory perspective. Therefore, we perform experiments to evaluate the influence of the proposed method when using different numbers of training samples in practice. As shown in Table~\ref{table:exp:ablation:sample-size}, we find that 1) the proposed method delivers  consistent improvements when using different numbers of training samples; and 2) increasing the number of training samples significantly improves the performance of heatmap regression models with low-resolution input images.

\textbf{The influence of different backbone networks}. If we do not take the heatmap prediction error into consideration, the quantization error in heatmap regression is then caused by the downsampling of heamaps: 1) the downsampling of input images and 2) the downsampling of CNN feature maps. Though the analysis of heamap prediction error is out the scope of this paper, we perform some experiments to demonstrate the influence of different feature maps from the backbone networks in practice. Specifically, we perform experiments using the following two settings: 1) upsampling the feature maps from HRNet~\cite{sun2019deep}; or 2) using the feature maps from U-shape backbone networks, i.e., U-Net~\cite{ronneberger2015u}. As shown in Table~\ref{table:appendix:backbone:wflw}, we see that 1) directly upsampling the feature maps achieves comparable performance with the baseline method; 2) U-Net performs better than HRNet-W18 when using a small input resolution (e.g., $64\times64$ pixels), while is significantly worse than HRNet when using a large input resolution (e.g., $256\times256$ pixels); and 3) U-Net contains more parameters and requires much more computations than HRNet when using the same input resolution. It would be interesting to further explore more efficient U-shape networks for low-resolution heatmap-based semantic landmark localization.

\begin{table}[ht]
\caption{The influence of different types of heatmap when using the heatmap regression model with different input resolutions.}
\label{table:appendix:heatmap}
\begin{center}
\begin{tabular}{|c|c|c|c|}
\hline
\multirow{2}{*}{Heatmap} & \multicolumn{3}{c|}{NME (\%)} \\
\cline{2-4} & $256\times256$ & $128\times128$ & $64\times64$ \\
\hline
Gaussian & 3.86 & 4.21 &  4.99 \\
\hline
Binary & 3.81 & 3.99 & 4.62  \\
\hline 
\end{tabular}
\end{center}
\end{table}

\begin{figure}[!ht]
\begin{center}
\centerline{\includegraphics[width=\linewidth]{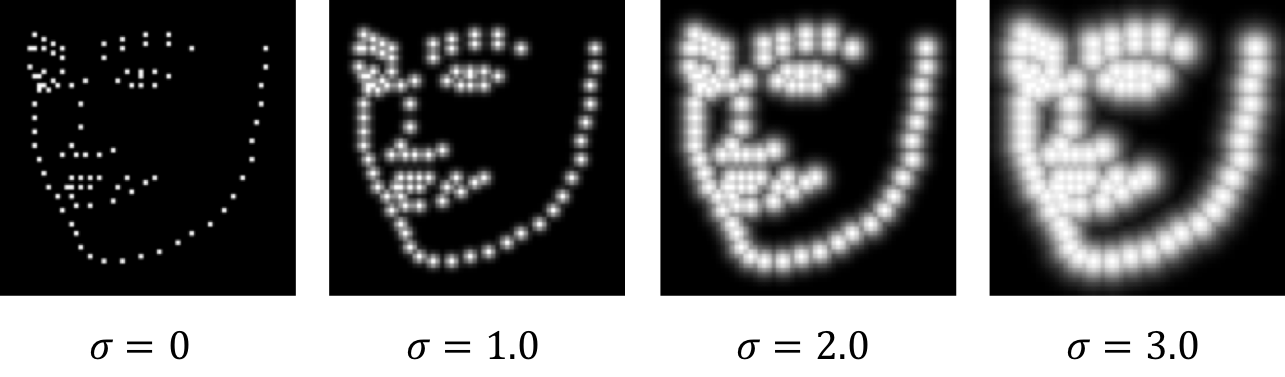}}
\end{center}
\caption{An intuitive example of the ground truth heatmap on WFLW dataset using different $\sigma$. We plot all heatmaps $\boldsymbol{h}^g_1, \dots, \boldsymbol{h}^g_{98}$ into a single figure for better visualization.}
\label{fig:appendix:heatmap}
\end{figure}

\textbf{The influence of different types of heatmap.} We perform some experiments to demonstrate the influence when using different types of heatmap, Gaussian heatmap and binary heatmap. As shown in Table~\ref{table:appendix:heatmap}, 1) when using a large input resolution, the heatmap regression model using either Gaussian heatmap or binary heatmap achieves comparable performance; and 2) when using a low input resolution, the heatmap regression model achieves better performance with the binary heatmap. We demonstrate the differences between binary heatmap and Gaussian heatmap in Figure~\ref{fig:appendix:heatmap}. Specifically, the Gaussian heatmap improves the robustness of heatmap prediction, while at the risk of increasing the uncertainty on the maximum activation point in the predicted heatmap. Therefore, when training very efficient heatmap regression models using a low input resolution, we recommend the binary heatmap.

\begin{figure*}[!ht]
\begin{center}
\centerline{\includegraphics[width=\linewidth]{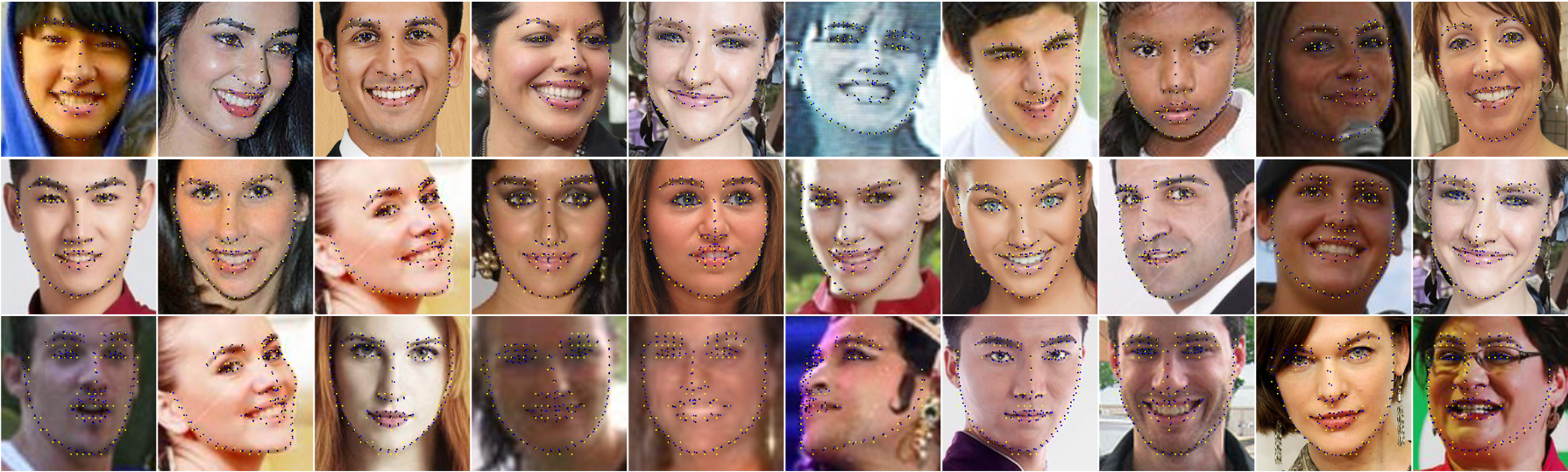}}
\end{center}
\caption{Qualitative results from the test split of the WFLW dataset (best view in color). \textbf{Blue}: the ground truth facial landmarks. \textbf{Yellow}: the predicted facial landmarks by the vanilla quantization system (i.e., $k=1$). \textbf{Black}: the predicted facial landmarks by the proposed quantization system (i.e., $k=9$). For above three rows, we use the heatmap regression models with the input resolutions, $256\times256$, $128\times128$, and $64\times64$ pixels, respectively.}
\label{fig:appendix:demo}
\end{figure*}

\textbf{The qualitative comparison between the vanilla quantization system and the proposed quantization system}. We provide some demo images for facial landmark detection using both the baseline method (i.e., $k=1$) and the proposed quantization method (i.e., $k=9$) to demonstrate the effectiveness of the proposed method for accurate semantic landmark localization. As shown in Fig.~\ref{fig:appendix:demo}, we see that the black landmarks are closer to the blue landmarks than the yellow landmarks, especially when using low resolution models (e.g., $64\times64$ pixels).

\subsection{Human Pose Estimation}

To utilize the proposed method for accurate semantic landmark localization, it contains only one hyper-parameter $k$, i.e., the number of activation points. To better understand the effectiveness of the proposed method for human pose estimation, we perform ablation studies using different numbers of alternative activation points on both MPII and COCO datasets. As shown in Table~\ref{table:exp:pose:coco:ablation} and Table~\ref{table:exp:pose:mpii:ablation}, we find that the proposed method achieves comparable performance when $k$ is between $10$ and $25$, making it easy to choose a proper $k$ on the validation data for human pose estimation applications.

\begin{table*}[ht]
\caption{Results on COCO validation set. The input resolution is $256\times192$ pixels. The first row indicates the performance when using the compensation method ``shift a quarter to the second maximum activation point".}
\label{table:exp:pose:coco:ablation}
\begin{center}
\begin{tabular}{|c|c|c|c|c|c|c|c|c|c|c|c|}
\hline
Backbone & $k$ & AP & Ap .5 & AP .75 & AP (M) & AP (L) & AR & AR .5 & AR .75 & AR (M) & AR (L) \\
\hline
HRNet-W32 & - & 0.744 & 0.905 & 0.819 & 0.708 & 0.810 & 0.798 & 0.942 & 0.865 & 0.757 & 0.858 \\
\hline
HRNet-W32 & 1 & 0.723 & 0.904 & 0.811 & 0.690 & 0.788 & 0.782 & 0.941 & 0.859 & 0.741 & 0.841 \\
\hline
HRNet-W32 & 2 &0.738 & 0.905 & 0.817 & 0.702 & 0.805 & 0.793 & 0.942 & 0.864 & 0.752 & 0.854 \\
\hline
HRNet-W32 & 3 & 0.743 & 0.905 & 0.819 & 0.708 & 0.810 & 0.797 & 0.942 & 0.865 & 0.756 & 0.857\\
\hline
HRNet-W32 & 4 & 0.743 & 0.904 & 0.819 & 0.709 & 0.809 & 0.797 & 0.941 & 0.866 & 0.756 & 0.857\\
\hline
HRNet-W32 & 5 & 0.747 & 0.904 & 0.819 & 0.712 & 0.814 & 0.800 & 0.940 & 0.866 & 0.759 & 0.859 \\
\hline
HRNet-W32 & 6 & 0.747 & 0.905 & 0.820 & 0.711 & 0.814 & 0.799 & 0.941 & 0.865 & 0.759 & 0.859\\
\hline
HRNet-W32 & 7 & 0.745 & 0.905 & 0.820 & 0.709 & 0.812 & 0.798 & 0.941 & 0.864 & 0.757 & 0.857\\
\hline
HRNet-W32 & 8 & 0.746 & 0.905 & 0.819 & 0.711 & 0.813 & 0.799 & 0.942 & 0.863 & 0.759 & 0.858\\
\hline
HRNet-W32 & 9 & 0.746 & 0.905 & 0.819 & 0.710 & 0.812 & 0.798 & 0.942 & 0.863 & 0.758 & 0.857\\
\hline
HRNet-W32 & 10 & 0.749 & 0.905 & 0.820 & 0.713 & 0.815 & 0.801 & 0.943 & 0.864 & 0.760 & 0.860 \\
\hline
HRNet-W32 & 11 &  \textbf{0.750} & \textbf{0.906} & \textbf{0.820} & \textbf{0.715} & \textbf{0.817} & \textbf{0.802} & \textbf{0.942} & \textbf{0.865} & \textbf{0.761} & \textbf{0.861} \\
\hline
HRNet-W32 & 12 &  0.750 & 0.906 & 0.821 & 0.714 & 0.817 & 0.801 & 0.942 & 0.865 & 0.760 & 0.861\\
\hline
HRNet-W32 & 13 &  0.749 & 0.906 & 0.821 & 0.713 & 0.816 & 0.800 & 0.942 & 0.865 & 0.760 & 0.860\\
\hline
HRNet-W32 & 14 &  0.749 & 0.905 & 0.820 & 0.713 & 0.816 & 0.800 & 0.941 & 0.864 & 0.760 & 0.860\\
\hline
HRNet-W32 & 15 &  0.749 & 0.906 & 0.820 & 0.713 & 0.817 & 0.800 & 0.942 & 0.863 & 0.760 & 0.860\\
\hline
HRNet-W32 & 16 &  0.749 & 0.905 & 0.820 & 0.713 & 0.816 & 0.800 & 0.941 & 0.863 & 0.760 & 0.860\\
\hline
HRNet-W32 & 17 &  0.750 & 0.905 & 0.821 & 0.714 & 0.817 & 0.801 & 0.942 & 0.863 & 0.761 & 0.860\\
\hline
HRNet-W32 & 18 &  0.750 & 0.905 & 0.820 & 0.713 & 0.817 & 0.801 & 0.942 & 0.864 & 0.760 & 0.860\\
\hline
HRNet-W32 & 19 &  0.750 & 0.904 & 0.820 & 0.714 & 0.816 & 0.801 & 0.941 & 0.864 & 0.760 & 0.860\\
\hline
HRNet-W32 & 20 & 0.749 & 0.904 & 0.820 & 0.713 & 0.816 & 0.800 & 0.940 & 0.864 & 0.760 & 0.859\\
\hline
HRNet-W32 & 21 &  0.747 & 0.904 & 0.819 & 0.712 & 0.813 & 0.799 & 0.940 & 0.862 & 0.758 & 0.858\\
\hline
HRNet-W32 & 22 &  0.749 & 0.904 & 0.819 & 0.713 & 0.815 & 0.799 & 0.941 & 0.863 & 0.759 & 0.859\\
\hline
HRNet-W32 & 23 &  0.749 & 0.904 & 0.819 & 0.713 & 0.816 & 0.800 & 0.941 & 0.863 & 0.760 & 0.860\\
\hline
HRNet-W32 & 24 &  0.749 & 0.904 & 0.820 & 0.713 & 0.817 & 0.800 & 0.941 & 0.863 & 0.759 & 0.860\\
\hline
HRNet-W32 & 25 & 0.749 & 0.905 & 0.820 & 0.713 & 0.817 & 0.800 & 0.941 & 0.864 & 0.760 & 0.860\\
\hline
HRNet-W32 & 30 & 0.749 & 0.905 & 0.819 & 0.714 & 0.818 & 0.800 & 0.941 & 0.862 & 0.759 & 0.859\\
\hline
HRNet-W32 & 35 & 0.748 & 0.902 & 0.819 & 0.712 & 0.816 & 0.799 & 0.940 & 0.862 & 0.758 & 0.859\\
\hline
HRNet-W32 & 40 & 0.746 & 0.901 & 0.818 & 0.710 & 0.814 & 0.797 & 0.939 & 0.861 & 0.756 & 0.857\\
\hline
HRNet-W32 & 45 & 0.746 & 0.901 & 0.815 & 0.710 & 0.814 & 0.796 & 0.939 & 0.860 & 0.755 & 0.857\\
\hline
HRNet-W32 & 50 & 0.746 & 0.901 & 0.814 & 0.710 & 0.816 & 0.797 & 0.938 & 0.859 & 0.755 & 0.857\\
\hline
HRNet-W32 & 75 & 0.741 & 0.899 & 0.811 & 0.705 & 0.810 & 0.792 & 0.936 & 0.856 & 0.751 & 0.853\\
\hline
HRNet-W32 & 100 & 0.734 & 0.897 & 0.805 & 0.697 & 0.803 & 0.785 & 0.934 & 0.850 & 0.743 & 0.846\\
\hline
HRNet-W32 & 125 & 0.717 & 0.893 & 0.795 & 0.682 & 0.785 & 0.770 & 0.930 & 0.841 & 0.727 & 0.831 \\
\hline 
HRNet-W32 & 150 & 0.651 & 0.889 & 0.755 & 0.629 & 0.702 & 0.713 & 0.925 & 0.809 & 0.680 & 0.762 \\
\hline
\end{tabular}
\end{center}
\end{table*}

\begin{table*}[ht]
\caption{Results on MPII validation set. The input resolution is $256\times256$ pixels. The first row indicates the performance when using the compensation method ``shift a quarter to the second maximum activation point".}
\label{table:exp:pose:mpii:ablation}
\begin{center}
\begin{tabular}{|c|c|c|c|c|c|c|c|c|c|c|}
\hline
Backbone & $k$ & Head & Shoulder & Elbow & Wrist &  Hip & Knee & Ankle & Mean & Mean@0.1  \\
\hline 
HRNet-W32 & - & 97.1 & 95.9 &  90.3 & 86.4 & 89.1 & 87.1 &  83.3 & 90.3 &  37.7 \\
\hline
HRNet-W32 & 1 & 97.033 & 95.703 & 90.131 & 86.312 & 88.402 & 86.802 & 82.924 & 90.055 & 32.808  \\
\hline 
HRNet-W32 & 2 & 97.033 & 95.856 & 90.302 & 86.416 & 88.818 & 87.004 & 83.325 & 90.247 & 35.912 \\
\hline
HRNet-W32 & 3 & 97.135 & 95.975 & 90.302 & 86.347 & 89.250 & 86.741 & 83.018 & 90.271 & 37.668 \\
\hline
HRNet-W32 & 4 & 97.237 & 95.907 & 90.643 & 86.141 & 89.164 & 86.862 & 83.089 & 90.294 & 36.997 \\
\hline
HRNet-W32 & 5 & 97.203 & 95.839 & 90.677 & 86.141 & 89.147 & 87.124 & 82.994 & 90.315 & 38.774 \\
\hline
HRNet-W32 & 6 & 97.101 & 95.839 & 90.677 & 86.262 & 89.181 & 86.923 & 83.089 & 90.310 & 38.548 \\
\hline
HRNet-W32 & 7 & 97.135 & 95.822 & 90.609 & 86.313 & 89.199 & 86.943 & 83.113 & 90.312 & 37.913 \\
\hline
HRNet-W32 & 8 & 97.033 & 95.822 & 90.557 & 86.175 & 89.147 & 86.922 & 82.876 & 90.245 & 38.129 \\
\hline
HRNet-W32 & 9 & 97.067 & 95.822 & 90.694 & 86.329 & 89.112 & 86.942 & 82.876 & 90.284 & 38.457 \\
\hline
HRNet-W32 & 10 & 97.033 & 95.873 & 90.626 & 86.141 & 89.372 & 86.862 & 82.995 & 90.297 &  39.024 \\
\hline
HRNet-W32 & 11 &  \textbf{97.101} &  \textbf{95.907} &  \textbf{90.592} &  \textbf{86.244} &  \textbf{89.406} &  \textbf{86.842} &  \textbf{82.853} &  \textbf{90.302} &  \textbf{39.139} \\
\hline
HRNet-W32 & 12 & 97.101 & 95.822 & 90.660 & 86.175 & 89.475 & 86.761 & 82.900 & 90.297 & 38.996 \\
\hline 
HRNet-W32 & 13 & 97.101 & 95.822 & 90.694 & 86.141 & 89.475 & 86.741 & 82.829 & 90.289 & 38.855 \\
\hline
HRNet-W32 & 14 & 97.033 & 95.822 & 90.728 & 86.141 & 89.337 & 86.640 & 82.806 & 90.250 & 38.616 \\
\hline
HRNet-W32 & 15 & 97.169 & 95.754 & 90.609 & 86.124 & 89.337 & 86.801 & 82.475 & 90.211 & 38.798 \\
\hline
HRNet-W32 & 16 & 97.101 & 95.771 & 90.694 & 86.141 & 89.199 & 86.882 & 82.569 & 90.226 & 38.691\\
\hline
HRNet-W32 & 17 & 97.067 & 95.822 & 90.609 & 86.142 & 89.285 & 86.821 & 82.617 & 90.224 & \textbf{39.053} \\
\hline
HRNet-W32 & 18 & 97.033 & 95.822 & 90.626 & 86.004 & 89.250 & 86.801 & 82.522 & 90.193 & \textbf{39.048}\\
\hline
HRNet-W32 & 19 & 97.033 & 95.839 & 90.626 & 85.953 & 89.250 & 86.822 & 82.593 & 90.198 & \textbf{39.058}\\
\hline
HRNet-W32 & 20 & 96.999 & 95.788 & 90.626 & 86.039 & 89.389 & 86.801 & 82.664 & 90.226 & 39.014\\
\hline
HRNet-W32 & 21 & 96.999 & 95.771 & 90.677 & 85.935 & 89.337 & 86.862 & 82.451 & 90.187 & 38.855 \\
\hline 
HRNet-W32 & 22 & 96.999 & 95.856 & 90.626 & 86.090 & 89.268 & 86.741 & 82.475 & 90.198 & 38.842\\
\hline 
HRNet-W32 & 23 & 96.999 & 95.890 & 90.609 & 86.038 & 89.164 & 86.801 & 82.499 & 90.187 & 39.037\\
\hline 
HRNet-W32 & 24 & 96.862 & 95.754 & 90.592 & 86.021 & 89.233 & 86.842 & 82.333 & 90.146 & 39.089\\
\hline 
HRNet-W32 & 25 & 96.930 & 95.754 & 90.506 & 86.055 & 89.302 & 86.721 & 82.309 & 90.135 & \textbf{39.126}\\
\hline
HRNet-W32 & 30 & 96.862 & 95.788 & 90.438 & 86.004 & 89.372 & 86.660 & 82.144 & 90.101 & 38.759 \\
\hline 
HRNet-W32 & 35 & 96.623 & 95.873 & 90.523 & 85.936 & 89.354 & 86.701 & 81.884 & 90.075 & 38.964\\
\hline 
HRNet-W32 & 40 & 96.385 & 95.856 & 90.523 & 85.816 & 89.406 & 86.721 & 81.908 & 90.049 & 38.618\\
\hline 
HRNet-W32 & 45 & 96.317 & 95.669 & 90.506 & 85.746 & 89.285 & 86.660 & 82.026 & 89.990 & 38.579\\
\hline  
HRNet-W32 & 50 & 96.214 & 95.686 & 90.506 & 85.678 & 89.268 & 86.439 & 81.648 & 89.899 & 38.584\\
\hline  
HRNet-W32 & 55 &  96.044 & 95.669 & 90.404 & 85.335 & 89.337 & 86.439 & 81.506 & 89.815 & 38.496\\
\hline  
HRNet-W32 & 60 & 95.805 & 95.533 & 90.302 & 85.147 & 89.147 & 86.419 & 81.270 & 89.672 & 38.332\\
\hline  
HRNet-W32 & 65 & 95.703 & 95.618 & 90.251 & 85.079 & 89.164 & 86.419 & 81.081 & 89.646 & 38.218\\
\hline  
HRNet-W32 & 70 & 95.498 & 95.584 & 90.165 & 85.215 & 89.302 & 86.378 & 80.963 & 89.633 & 38.197\\
\hline  
HRNet-W32 & 75 & 95.259 & 95.533 & 90.029 & 85.147 & 89.147 & 86.318 & 80.467 & 89.487 & 38.054\\
\hline  
HRNet-W32 & 80 & 95.020 & 95.516 & 90.046 & 84.907 & 89.216 & 86.197 & 80.255 & 89.404 & 37.684\\
\hline  
HRNet-W32 & 85 & 94.679 & 95.465 & 89.927 & 84.925 & 89.337 & 86.258 & 80.042 & 89.357 & 37.463\\
\hline  
HRNet-W32 & 90 & 94.407 & 95.431 & 89.790 & 84.907 & 89.233 & 86.197 & 79.735 & 89.251 & 37.208 \\
\hline  
HRNet-W32 & 95 & 94.065 & 95.414 & 89.773 & 84.890 & 89.302 & 86.097 & 79.263 & 89.165 & 36.719\\
\hline  
HRNet-W32 & 100 & 93.827 & 95.482 & 89.603 & 84.737 & 89.320 & 86.076 & 78.578 & 89.032 & 36.180\\
\hline  
HRNet-W32 & 105 & 93.520 & 95.448 & 89.603 & 84.599 & 89.199 & 85.996 & 77.964 & 88.884 & 35.532\\
\hline
HRNet-W32 & 110 & 93.008 & 95.448 & 89.518 & 84.273 & 89.147 & 85.835 & 77.114 & 88.657 & 34.770\\
\hline
HRNet-W32 & 115 & 92.565 & 95.296 & 89.501 & 84.085 & 89.112 & 85.794 & 76.405 & 88.483 & 33.352\\
\hline
HRNet-W32 & 120 & 92.121 & 95.262 & 89.398 & 83.930 & 89.060 & 85.533 & 75.272 & 88.238 & 31.814\\
\hline
HRNet-W32 & 125 & 91.473 & 95.160 & 89.160 & 83.554 & 88.991 & 85.472 & 74.162 & 87.939 & 30.065\\
\hline
HRNet-W32 & 130 & 90.825 & 94.939 & 89.211 & 83.023 & 89.095 & 85.089 & 72.910 & 87.609 & 27.528\\
\hline
HRNet-W32 & 135 & 90.246 & 94.667 & 89.006 & 82.577 & 88.852 & 84.807 & 71.209 & 87.164 & 24.757\\
\hline
HRNet-W32 & 140 & 89.529 & 94.463 & 88.853 & 82.046 & 88.679 & 83.800 & 69.792 & 86.659 & 21.788\\
\hline
HRNet-W32 & 145 & 88.404 & 94.124 & 88.717 & 81.310 & 88.645 & 82.773 & 67.997 & 86.053 & 19.136\\
\hline
HRNet-W32 & 150 & 87.756 & 93.886 & 88.614 & 80.146 & 88.575 & 81.383 & 66.084 & 85.373 & 16.586\\
\hline
HRNet-W32 & 175 & 82.401 & 90.829 & 86.927 & 71.515 & 86.014 & 70.504 & 56.826 & 80.109 & 9.204\\
\hline
HRNet-W32 & 200 & 68.008 & 87.075 & 82.325 & 62.146 & 79.176 & 58.737 & 41.120 & 72.009 & 5.988\\
\hline
\end{tabular}
\end{center}
\end{table*}

\end{document}